\documentclass[sigconf]{acmart}

\usepackage{algorithm}
\usepackage{algpseudocode}
\usepackage{xcolor}
\usepackage{colortbl}
\usepackage{booktabs}
\usepackage{makecell}
\usepackage{multirow}
\usepackage{siunitx}
\usepackage{threeparttable}
\usepackage{tabularx}
\usepackage{pifont}
\usepackage{pdfrender}
\usepackage{subcaption}
\newcommand{\best}{%
  \pdfrender{TextRenderingMode=FillStroke,LineWidth=0.1pt}%
  \bfseries
}
\usepackage{comment}
\newcommand{\cmark}{\textcolor{green!60!black}{\ding{51}}}
\newcommand{\xmark}{\textcolor{red!70!black}{\ding{55}}}

\newcommand{\cmarkbold}{\textcolor{green!60!black}{\textbf{\ding{51}}}}

\newlength{\statbarlen}
\definecolor{statbarfill}{RGB}{86,118,176}
\newcommand{\statbar}[1]{%
  \hbox{%
    \rlap{\textcolor{black!9}{\rule{\statbarlen}{4.4pt}}}%
    \rlap{\textcolor{statbarfill}{\rule{#1\statbarlen}{4.4pt}}}%
    \hspace{\statbarlen}%
  }%
}
\newcommand{\statpanel}[1]{%
  \multicolumn{4}{@{}l}{\cellcolor{black!6}\textsc{#1}} \\
}
\newcommand{\statcnt}[1]{\makebox[1.7em][r]{#1}}
\newcommand{\statpct}[1]{\makebox[2.6em][r]{#1\%}}
\newcommand{\statrow}[3]{%
  \statcnt{#1}\,\statpct{#2}\hspace{3pt}\statbar{#3}%
}

\AtBeginDocument{%
  }

\setcopyright{acmlicensed}
\copyrightyear{2018}
\acmYear{2018}
\acmDOI{XXXXXXX.XXXXXXX}
\acmConference[Conference acronym 'XX]{Make sure to enter the correct
  conference title from your rights confirmation email}{June 03--05,
  2018}{Woodstock, NY}
\acmISBN{978-1-4503-XXXX-X/2018/06}

\begin{document}

\title[RSIV-Harness]{RSIV-Harness: A Recursive Self-Improvement Harness Framework for Open-Domain Conversational Text-to-Visualization}

\author{Jiawei Lu$^{1}$, Chen Zhang$^{1}$, Jinwei Lu$^{1}$, Yuanwei Song$^{1}$, Yuanfeng Song, Raymond Chi-Wing Wong$^{2}$}

\affiliation[obeypunctuation=true]{%
  \institution{$^{1}$The Hong Kong Polytechnic University, Hong Kong, China\\
  $^{2}$The Hong Kong University of Science and Technology, Hong Kong, China}%
  \country{\mbox{}}%
}

\begin{abstract}
Users often request charts without knowing which database or table holds the relevant data.
Such requests may be ambiguous, and users may ask follow-up questions to deepen their analysis or iteratively refine existing charts.
A useful assistant must therefore discover relevant data sources among heterogeneous databases, ask clarifying questions when needed, and track evolving user intent while preserving constraints established in earlier turns.
Yet recent text-to-visualization agents typically rely on static agent harnesses that are predefined and remain unchanged despite recurring failures across tasks.
To address this limitation, we introduce \textsc{RSIV-Harness}, which keeps
the language model weights fixed and applies \textbf{R}ecursive
\textbf{S}elf-\textbf{I}mprovement to revise the surrounding harness
from the agent's own failure traces.
To balance flexible self-improvement with stability,
we introduce a four-layer framework that evaluates proposed harness revisions through anchor and memory checks,
refining those that pass and restoring the best validated harness otherwise.
We further introduce \textsc{ODAC-Vis}, a benchmark with 640 tasks for \textbf{O}pen-\textbf{D}omain, \textbf{A}mbiguous, and \textbf{C}onversational text-to-visualization
that jointly evaluates open-domain data discovery, ambiguity resolution, multi-turn visualization refinement, and executable chart generation.
Experiments show that a single round of \textsc{RSIV-Harness} improves all six evaluation metrics relative to the seed harness,
with particularly large gains in executability (+28.02 points) and chart structure (+34.01 points),
demonstrating the ability to learn from task-solving experience and turn recurring failures across tasks into substantial performance improvements.
\end{abstract}



\maketitle

\section{Introduction}

\begin{figure*}[t]
  \centering
  \includegraphics[width=\textwidth]{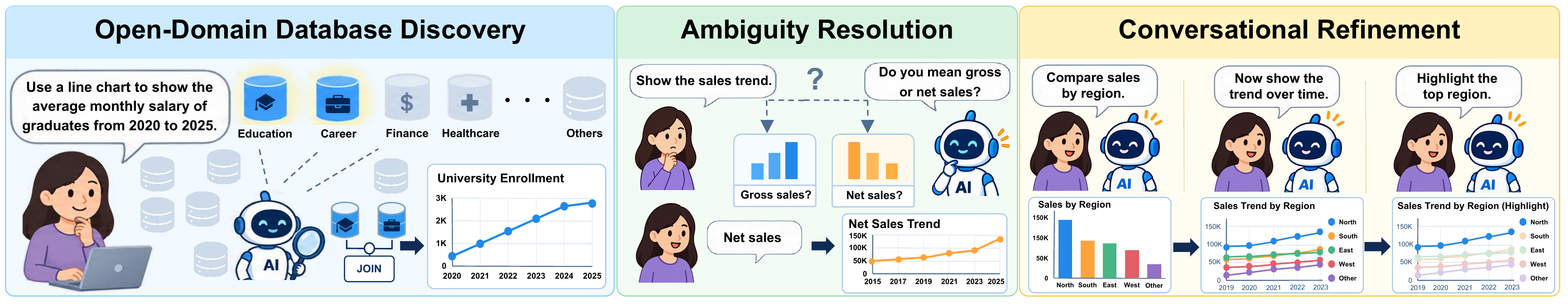}
  \caption{Three challenges in real-world
    text-to-visualization.
    The assistant agent needs to discover relevant databases, resolve ambiguous
    requests before charting, and refine the chart across dialogue
    turns.}
  \Description{Three side-by-side scenarios: discovering databases for
    a salary chart, clarifying gross versus net sales, and refining a
    regional sales chart over three dialogue turns.}
  \label{fig:motivation}
\end{figure*}

While extensive research on natural language interfaces for
automated visualization within the database and data
mining community~\cite{luo2021nvbench,luo2024viseval,yu2019sparc,yu2019cosql,deng2026taco,ouyang2025nvagent,lu2026multivis,abacussql2025},
recent agent systems increasingly wrap large language models in an
\emph{agent harness}~\cite{lin2026ahe,lee2026meta,liu2026adaptive}:
a surrounding layer of system rules, tool interfaces, and control
policies that govern schema access, SQL generation, and chart
rendering.
Industrial and academic agents~\cite{talaei2024chess, lu2026demonstration}
show that such harnesses can coordinate retrieval, query planning,
execution feedback, and visualization.
Yet most harnesses are authored once and kept fixed after deployment,
while public benchmarks still largely assume a known target database
and a well specified single-turn
request~\cite{luo2021nvbench,luo2024viseval,yu2019sparc}.
These gaps limit deployment in enterprise data lakes and rigorous
measurement of open-domain conversational text-to-visualization.

\textbf{Benchmark and task coverage.}
In enterprise data lakes, analysts often begin without a known target
database, must resolve ambiguous requests through dialogue, and iterate
on SQL and charts over multiple turns before trusting the output (Figure~\ref{fig:motivation}).
Existing benchmarks address these needs only in isolation: open-domain
and cross-database text-to-SQL~\cite{deng2026taco,abacussql2025},
controlled ambiguity in conversational querying~\cite{clarity2025,nvbench22025},
and multi-turn text-to-visualization~\cite{luo2021nvbench,luo2024viseval,wong2024dialnvbench,text2vis2025}.
To our knowledge, no public benchmark integrates them for end-to-end
evaluation with executable query and chart supervision throughout the
dialogue.

\textbf{Deployment and adaptation.}
State-of-the-art visualization agents improve reliability through
fixed multi-stage workflows or within-task execution
feedback~\cite{ouyang2025nvagent,lu2026multivis}.
Three adaptation problems therefore remain.
First, recurring failure modes such as premature code generation, schema search misuse, and cross-database boundary violations persist across tasks because the harness is not revised from accumulated traces.
Second, improving reliability often requires retraining or replacing the base model rather than evolving the surrounding harness under fixed weights.
Third, learning from failed tasks may introduce regressions, causing the harness to fail on tasks it previously solved correctly.

\textbf{Our solution.}
We introduce \textsc{ODAC-Vis} and \textsc{RSIV-Harness} as a co-designed
benchmark and method response.
\textsc{RSIV-Harness} keeps the language model weights fixed.
It revises only the surrounding harness: the system surrounding the
model, including its prompt, tool, memory, skill, and middleware.
At each turn, an inner loop either clarifies the request or generates a
visualization program, and retries with observed errors
within a bounded budget.
An outer loop evaluates failure traces on a stratified training sample
and proposes a revision of the harness.
Rather than following predefined edit rules, the outer loop autonomously determines
which harness components to revise and how, guided by observed failure patterns.
To prevent such open-ended evolution from degrading previously acquired capabilities,
we introduce a four-layer stability framework that accepts revisions only when
they satisfy both validation-anchor and regression-memory checks,
reverting to the best validated harness otherwise.
We establish formal guarantees that accepted revisions refresh protected records,
rejected revisions preserve those records and restore the champion,
the best validation score remains nondecreasing,
and regression on sampled protected records stays bounded.
To evaluate this experience-driven evolution under realistic open-domain conversational visualization challenges,
we further introduce \textsc{ODAC-Vis}, a benchmark for \textbf{O}pen \textbf{D}omain,
\textbf{A}mbiguous, and \textbf{C}onversational text-to-visualization.
\textsc{ODAC-Vis} jointly evaluates source discovery across a fully exposed database collection,
ambiguity resolution, and multi-turn visualization refinement,
with domain-held-out evaluation and executable query-and-chart supervision at each code-generating turn.
Together, \textsc{ODAC-Vis} provides a unified testbed for these interconnected challenges,
while \textsc{RSIV-Harness} enables agents to learn from recurring task failures and improve their behavior across tasks.

\textbf{Contributions.}
Our work makes four main contributions:
\begin{itemize}
\item \textbf{ODAC-Vis Benchmark.} We present the first
benchmark in which agents must generate visualizations without being given the target database
or table, clarify ambiguous requests, and support follow-up refinement.
\item \textbf{RSIV-Harness Method.} We propose a recursive self improvement framework
that enables agents to evolve from their own task-solving experience.
An inner ReAct loop handles conversational visualization tasks,
while an outer loop evaluates execution traces,
identifies recurring failures across tasks,
and autonomously proposes targeted revisions to the surrounding harness.
\item \textbf{Stability Rules \& Formal Guarantees.} We develop a four-layer stability rule framework
to ensure stable evolution without restricting how harness revisions are proposed.
We formally guarantee protected-memory updates upon acceptance, champion restoration without memory modification upon rejection,
nondecreasing champion validation scores, and bounded regression on sampled protected records.
\item \textbf{Empirical Study.} Extensive experiments on held-out domains
show that a single round of recursive self-improvement improves all six evaluation metrics
over the seed harness. Cross-backbone transfer and ablation studies further examine
the generality of these improvements and the contributions of individual components.
These findings demonstrate that agents can turn accumulated task failures into substantial improvements on previously unseen tasks.
\end{itemize}

\section{Related Work}
\label{sec:related}

This work is closely related to conversational text-to-visualization
and to recursive self-improvement of agent harnesses.
In this section, we review the prior literature in these two main
directions.

\subsection{Text-to-Visualization and Conversational Data Access}
Natural language (NL) interfaces for visualization (Vis) have progressed from
direct code generation~\cite{maddigan2023chat2vis} to curated
benchmarks and agentic pipelines.
Early text-to-vis datasets such as nvBench and
VisEval~\cite{luo2021nvbench,luo2024viseval} provide large-scale
NL-Vis pairs, enabling automatic
validity and readability checks but not source discovery over a full
database collection.
Subsequent work extends coverage along three largely separate axes.
Ambiguity-aware benchmarks such as nvBench~2.0 and
CLARITY~\cite{nvbench22025,clarity2025} inject lexical, schema, or
value ambiguity into querying, yet still assume a specified schema and
stop at SQL execution.
Complementary models improve robustness to lexical and phrasal
variability~\cite{lu2025lexical} and to ambiguity in interactive
text-to-visualization~\cite{lu2026robustvis,xu2026visinteract}.
Multi-turn visualization datasets including Dial-NVBench, Text2Vis, and
PMVisBench~\cite{wong2024dialnvbench,text2vis2025,wen2026pmvis}
support conversational refinement or progressive intent elaboration,
but provide fixed table or schema context rather than open-domain
routing across heterogeneous sources.
CoVis studies neural and LLM-driven multi-turn interactions for
conversational text-to-visualization~\cite{song2026covis}, and
FeVisQA studies free-form question answering over data
visualizations~\cite{song2025fevisqa}.
On the querying side, cross-domain semantic parsing benchmarks such as
Spider, BIRD, SParC, and CoSQL~\cite{yu2018spider,li2023bird,yu2019sparc,yu2019cosql}
establish standard splits for schema generalization and multi-turn SQL
dialogue, while TACO~\cite{deng2026taco} moves toward unspecified
databases and cross-database SQL composition.
Text-to-NoSQL translation extends natural-language querying to
non-relational stores~\cite{lu2026texttonosql}.
None of these resources jointly evaluate open-domain source discovery,
typed ambiguity resolution, executable query and chart
supervision, and multi-turn conversational interactions, which motivates our \textsc{ODAC-Vis} benchmark.

Visualization agents improve reliability through staged workflows and
within-task validation rather than cross-task adaptation.
nvAgent decomposes processing, composition, and validation in a fixed
collaborative workflow~\cite{ouyang2025nvagent}, and MultiVis-Agent
adds logic rules for bounded tool use and cross-modal
refinement~\cite{lu2026multivis,lu2026demonstration}.
Open-domain SQL systems such as Abacus-SQL combine retrieval with
multi-turn query generation~\cite{abacussql2025}, and recent agentic
NL2SQL frameworks employ multi-agent decomposition, rewriting, and
execution-grounded repair~\cite{talaei2024chess,wang2025macsql,pourreza2023dinsql,cen2025sqlfixagent,bogdanov2026agentnlq,luo2026qbridge,chen2026gbvsql}.
These systems nonetheless ship with prompts, tools, and control
policies fixed before deployment and do not revise the surrounding
harness from recurring failures observed across tasks.

\subsection{Agent Harnesses and Recursive Self-Improvement}
LLM agents are increasingly mediated by a surrounding \emph{harness}:
system rules, tool interfaces, middleware, and control policies that
govern how a frozen model interacts with an
environment~\cite{lin2026ahe,lee2026meta,liu2026adaptive}.
Industrial and academic text-to-SQL agents instantiate such harnesses
through schema retrieval, query planning, execution feedback, and
validator loops~\cite{talaei2024chess,wang2025macsql,ouyang2025nvagent,lu2026multivis},
yet most are hand engineered once and left unchanged after release.

Recursive self-improvement has recently been studied along three
complementary lines.
Weight-centric RSI agents such as AREX~\cite{arex2026} alternate an
inner task loop with an outer audit-and-refine loop, but sustain gains
through agentic mid-training and reinforcement learning over model
parameters rather than deployment-time harness edits.
Harness-evolution methods keep the base model frozen and instead
revise prompts, tools, and control programs from execution
traces~\cite{zhang2026selfharness,nie2026tthe,guo2026drevo,lin2026ahe,zheng2026seagym},
including test-time program search, weakness-driven proposal with
regression testing, and observability-driven component editing; yet
they neither support clarification over open database
collections nor regression-validated harness commits for coupled
query-and-chart execution in multi-turn dialogue.
Inference-time self-correction methods such as Reflexion, Self-Refine,
and DSPy~\cite{shinn2023reflexion,madaan2023selfrefine,khattab2024dspy}
improve individual trajectories through verbal feedback, iterative
drafting, or compiled prompt pipelines, but do not commit persistent
harness revisions validated across a task distribution.
\textsc{RSIV-Harness} addresses these gaps by learning from recurring failures
across tasks to autonomously evolve the agent harness,
while using regression-aware checks to preserve previously successful behaviors.

\section{The \textsc{ODAC-Vis} Benchmark}
\label{sec:odac}

\subsection{Overview and Motivation}
\label{sec:odac-overview}

We introduce \textsc{ODAC-Vis}, a benchmark for \textbf{O}pen
\textbf{D}omain, \textbf{A}mbiguous, and \textbf{C}onversational
text-to-visualization.
Each task simulates an analyst who requests a chart in natural language
without naming a target database, may need to clarify underspecified
intent, and may refine the analysis or the chart over subsequent turns.
The benchmark exposes all \emph{206} SQLite databases from the Spider
test collection~\cite{yu2018spider} and requires a model to discover
relevant sources, compose SQL when multiple databases are needed, and
emit executable Python that queries data and renders a Plotly chart.

\textbf{Open domain and cross-domain.}
Unlike prior text-to-vis datasets that bind each instance to a known
table or schema~\cite{luo2021nvbench,luo2024viseval,wong2024dialnvbench,text2vis2025},
\textsc{ODAC-Vis} treats the full database collection as visible at
inference time while keeping the gold target as one or more specific
databases per task.
Models must route to the correct sources without being told which
database to open, and cross-database tasks require SQL-only composition
through \texttt{ATTACH}, temporary tables, joins, semi-joins, unions,
or scalar subqueries rather than post-hoc Python merges.
Evaluation uses a \emph{domain-held-out} split: domains seen during
training never appear in validation or test, so success requires
generalization rather than memorizing a fixed schema catalog.
Complexity is stratified into single-table (30\%), single-database
join (50\%), and cross-database (20\%) tiers within each split.

\textbf{Ambiguity.}
Each task contains one to three ambiguity points drawn from eight
template-defined types.
\emph{Data semantics:}
\begin{itemize}
  \item \emph{Column selection}: ``Show revenue'' when both gross and
    net sales columns are valid measures.
  \item \emph{Aggregation granularity}: ``Show the trend'' when the
    request can be grouped by month or by quarter.
  \item \emph{Filter logic}: ``Top performers'' when ranking by revenue
    or by profit, or when conjunctive and disjunctive filters differ.
  \item \emph{Unit scale}: ``Plot sales'' when values may be shown in
    raw units, millions, or percentages.
\end{itemize}
\emph{Visualization semantics:}
\begin{itemize}
  \item \emph{Chart type}: ``Visualize growth'' when a line chart
    (trend) and a bar chart (comparison) are both reasonable.
  \item \emph{Encoding mapping}: ``Highlight categories'' when color
    or point size could encode the same dimension.
  \item \emph{Axis label}: ``Show enrollment'' when the y-axis may
    label counts or normalized rates.
  \item \emph{Interactivity}: ``Make it filterable'' when filtering by
    dropdown selection or brush interaction is intended.
\end{itemize}
Every ambiguity is backed by at least two schema-supported alternatives
that yield distinguishable query or chart outcomes, and each is resolved
through a clarification question followed by a user reply
that uniquely selects one alternative.
Dialogues never expose database or table names during clarification.

\textbf{Conversational structure.}
Tasks follow one of two scenarios, which set the analytical role of
the dialogue.
In \emph{drilldown}, later requests deepen the current analysis.
In \emph{iterate}, later requests revise the current chart.
Ambiguity is placed independently of that role.
Each task contains one to three ambiguity points on randomly chosen
request turns other than the final turn, so clarification may open
the dialogue or arise during a later drilldown or iteration.
A turn that contains an ambiguity requires the model to ask the user a
clarification question on that turn.
On the next turn the user answers the question, and the model writes
the visualization code for user's intent.
A turn without ambiguity requires the model to write the visualization
code directly.
Conversations are capped at eight user--assistant exchanges.
The two scenarios appear in equal proportion within each split.

\subsection{Dataset Construction Methodology}
\label{sec:odac-construction}

\textsc{ODAC-Vis} was constructed through a principled, LLM-assisted pipeline
with expert validation, prioritizing quality, coverage, and
faithfulness in open-domain, ambiguity-aware, multi-turn conversational
text-to-visualization scenarios. The detailed construction
process of \textsc{ODAC-Vis} is described below:

\textbf{Step 1: Cross-domain split.}
Spider provides 206 independent SQLite databases and 139
domains.
Domains are assigned to splits as atomic units: roughly 15 domains are
held out exclusively for test, while the remaining domains supply train
and validation tasks in a 5:1 ratio.
Within every split we enforce independent quotas over the Cartesian
product of scenario and complexity tier, so evaluation is not skewed
toward a single dialogue pattern or query difficulty.
All databases referenced by a cross-database task belong to domains in
the same split.
At inference time the agent may access the full database collection
$\mathcal{D}=\{d_1,\ldots,d_{206}\}$, while task gold specifies a target
subset $B^\star(t)\subseteq\mathcal{D}$ without naming it in user
dialogue; held-out test domains never appear in train or validation.
The released benchmark contains $N{=}640$ tasks.
Each construction slot is indexed by
$u=(s,\sigma,\tau)\in\mathsf{S}$, where $s\in\mathcal{S}$ is the
data split, $\sigma\in\Sigma$ the conversational scenario, and
$\tau\in\mathcal{T}$ the query-complexity tier, with
\begin{equation}
\label{eq:slot-space}
\mathsf{S}
=
\mathcal{S}\times\Sigma\times\mathcal{T},
\qquad
|\mathsf{S}|
=
|\mathcal{S}|\cdot|\Sigma|\cdot|\mathcal{T}|.
\end{equation}
We target the following marginal proportions over these dimensions:
train:validation:test in $14{:}3{:}3$ (approximately 70:15:15, yielding
448/96/96 tasks at $N{=}640$), equal weights on the two scenarios in
$\Sigma$, and complexity tiers in $\mathcal{T}$ in a global 3:5:2 ratio
(30/50/20\% single-table, single-database join, and cross-database).
Slot counts are obtained by nested proportional rounding:
\begin{equation}
\label{eq:quota-grid}
Q[s,\sigma,\tau]
=
\mathrm{Alloc}\!\left(
  \mathrm{Alloc}\!\left(
    \mathrm{Alloc}(N,\; r_{\mathcal{S}})[s],\; r_{\Sigma}
  \right)[\sigma],\; r_{\mathcal{T}}
\right)[\tau],
\end{equation}
where $Q[s,\sigma,\tau]$ is the number of tasks allocated to slot
$(s,\sigma,\tau)$, $r_{\mathcal{S}}$, $r_{\Sigma}$, and $r_{\mathcal{T}}$
are non-negative integer weight vectors over splits, scenarios, and
tiers, and $\mathrm{Alloc}(n,r)$ maps a total~$n$ and weights~$r$ to
integer counts that sum to~$n$ by largest-remainder rounding.
Applying Eq.~\eqref{eq:quota-grid} top-down first fixes the split
sizes, then balances scenarios within each split, and finally assigns
complexity tiers within each scenario cell, so every marginal target is
met without favoring a single dialogue pattern or query difficulty.

\begin{table*}[t]
  \centering
  \caption{Comparison of \textsc{ODAC-Vis} with existing benchmarks.}
  \label{tab:benchmark-comparison}
  \fontsize{9pt}{11.5pt}\selectfont
  \renewcommand{\arraystretch}{1.10}
  \setlength{\tabcolsep}{4pt}
  \begin{threeparttable}
  \begin{tabularx}{\textwidth}{@{}l c c c *{5}{>{\centering\arraybackslash}X}@{}}
  \toprule
  \textbf{Dataset} &
  \textbf{Scale} &
  \textbf{Domain} &
  \textbf{\makecell[c]{Target}} &
  \textbf{\makecell[c]{Open-\\domain}} &
  \textbf{\makecell[c]{Cross-\\domain}} &
  \textbf{\makecell[c]{Multi-turn\\dialogue}} &
  \textbf{\makecell[c]{Typed\\ambiguity}} &
  \textbf{\makecell[c]{Cross-\\database\tnote{a}}} \\
  \midrule
  Text2Vis~\cite{text2vis2025} &
    1985 NLQs &
    103 &
    Python &
    \xmark & \xmark & \cmark & \cmark &
    \xmark \\
  nvBench~2.0~\cite{nvbench22025} &
    7{,}878 NLQs &
    153 &
    Vega-Lite &
    \xmark & \xmark & \xmark & \cmark &
    \xmark \\
  TACO~\cite{deng2026taco} &
    14{,}500 NLQs &
    31 &
    SQL &
    \cmark & \xmark & \xmark & \cmark &
    \cmark \\
  Dial-NVBench~\cite{wong2024dialnvbench} &
    7{,}244 dialogues &
    105 &
    VQL &
    \xmark & \xmark & \cmark & \xmark &
    \xmark \\
  CoSQL~\cite{yu2019cosql} &
    3{,}007 dialogues &
    138 &
    SQL &
    \xmark & \cmark & \cmark & \cmark &
    \xmark \\
  MultiVis-Bench~\cite{lu2026multivis} &
    1{,}202 NLQs &
    105 &
    Python &
    \xmark & \xmark & \xmark & \xmark &
    \xmark \\
  \midrule
  \rowcolor{blue!5}
  \textbf{\textsc{ODAC-Vis}} &
    \textbf{640 dialogues} &
    \textbf{139} &
    \textbf{Python} &
    \cmarkbold & \cmarkbold & \cmarkbold & \cmarkbold &
    \cmarkbold \\
  \bottomrule
  \end{tabularx}
  \begin{tablenotes}
  \footnotesize
  \item[a] \textbf{Cross-database:} whether tasks require composing
  evidence from more than one database through executable SQL.
  \end{tablenotes}
  \end{threeparttable}
\end{table*}

\textbf{Step 2: Cross-database tasks.}
About 20\% of tasks require evidence from more than one Spider
database, yet the collection provides no predefined links between
files.
We model the collection as a composability graph
$G=(\mathcal{D},\Pi)$ whose nodes are databases and whose edges are
composition plans $\pi=(\rho,B,\kappa)$, where $\rho$ is a SQL
relation type, $B\subseteq\mathcal{D}$ the participating databases, and
$\kappa$ an executable program skeleton~\cite{deng2026taco}.
Candidate plans are generated from schema overlap and then filtered in
three stages.
First, relation-specific generators produce a candidate set
\begin{equation}
\label{eq:pi-candidates}
\Pi^{(0)}
=
\bigcup_{\rho\in\mathcal{R}}
\Pi^{(0)}_{\rho},
\qquad
\mathcal{R}
=
\{\textsc{equi\_join},\textsc{semi\_join},\textsc{union},\textsc{subq}\}.
\end{equation}
Equi-join candidates are retained only if they pass a staged quality
test on name similarity, value overlap, coverage, key uniqueness, and
join expansion:
\begin{equation}
\label{eq:equi-quality}
\Pi_{\textsc{equi\_join}}
=
\bigcup_{\theta\in\Theta}
\Bigl\{
  \pi\in\Pi^{(0)}_{\textsc{equi\_join}}
  \;\Big|\;
  \mathrm{Quality}(\pi,\theta)=1
\Bigr\},
\end{equation}
where $\Theta=\{\text{strict},\text{relaxed}_1,\text{relaxed}_2\}$
denotes progressively relaxed thresholds.
Next, every candidate must survive an execution probe that materializes
the skeleton on the reserved databases:
\begin{equation}
\label{eq:probe}
\Pi
=
\Bigl\{
  \pi\in\Pi^{(0)}
  \;\Big|\;
  \mathrm{Probe}(\kappa,B)\neq\bot
\Bigr\}.
\end{equation}
Finally, only plans whose databases lie in the same split are eligible
for construction in that split:
\begin{equation}
\label{eq:pi-split}
\Pi_s
=
\Bigl\{
  \pi\in\Pi
  \;\Big|\;
  B(\pi)\subseteq\mathcal{D}_s
\Bigr\},
\end{equation}
\begin{equation}
\label{eq:crossdb-capacity}
|\Pi_s|
\;\ge\;
\alpha\sum_{\sigma\in\Sigma} Q[s,\sigma,\textsc{cross\_db}],
\end{equation}
with $\mathcal{D}_s$ the databases reachable from domains in split~$s$
and $\alpha{=}3$ a safety margin used during split planning.
Cross-database gold is sampled as $\pi\sim\mathrm{Unif}(\Pi_s)$ and realized
in SQL via \texttt{ATTACH}; post-query pandas transforms are restricted
to unary cleaning on the final result set.

This keeps cross-database gold faithful to real schema overlap rather
than hand-written Python merges, and it aligns the benchmark with the
SQL-only composition protocol used in open-domain text-to-SQL
evaluation.

\textbf{Step 3: Task generation.}
We build tasks with a chart-centric hybrid of templates and LLMs.
Templates sample query intents, ambiguity points, and scenario-specific
follow-ups from schema metadata, compile each sample into structured
query, cleaning, and chart specifications together with executable SQL
and Plotly code, and discard instances that fail execution or ambiguity
checks.
An LLM (DeepSeek-V4.1-Flash~\cite{deepseekai2026deepseekv41flash}) then paraphrases the verified slots into natural dialogue.
For \emph{drilldown}, templates also sample a follow-up
analysis path after the initial chart; for \emph{iterate},
the target chart and iteration request are fixed first and clarification
turns are inserted afterward.
Each task contains at most eight user--assistant turns, and ambiguity
points are injected at randomly chosen request-bearing turns.
Dialogue length and ambiguity count are coupled so that one ambiguity
uses three or four turns, two use five or six, and three use seven or
eight; adjacent ambiguity turns are separated by at least one
non-ambiguous round:
\begin{equation}
\label{eq:ambiguity-schedule}
\mathcal{A}(T,k)
=
\Bigl\{
  \{a_1,\ldots,a_k\}
  \subseteq\{1,\ldots,T-1\}
  \;\Big|\;
  a_{i+1}-a_i \ge 2
\Bigr\}.
\end{equation}
For each ambiguity turn, templates construct at least two
schema-supported alternatives whose raw-result, cleaned-result, and
chart fingerprints are pairwise distinct, so clarification changes the
downstream gold rather than surface wording alone.
Global occurrence targets over the eight ambiguity types are matched to
task schedules by a unit-capacity bipartite max-flow assignment that
fixes one primary type per task and forbids repeating a type within a
dialogue.
A final validator rejects dialogues that violate the clarify-and-code
grammar in Section~\ref{sec:odac-overview}.

\textbf{Step 4: Gold artifacts and validation.}
Every \texttt{code} turn is released with a structured artifact that
links natural language to its specifications, executable program, and
reference chart, while each ambiguity point retains all viable
alternatives and the resolved path.
Duplicate trajectories are removed and an expert
check confirms overall dataset quality before release.
These annotations ground the evaluation protocols in
Section~\ref{sec:experiments}.

\begin{figure*}[t]
  \centering
  \includegraphics[width=\textwidth]{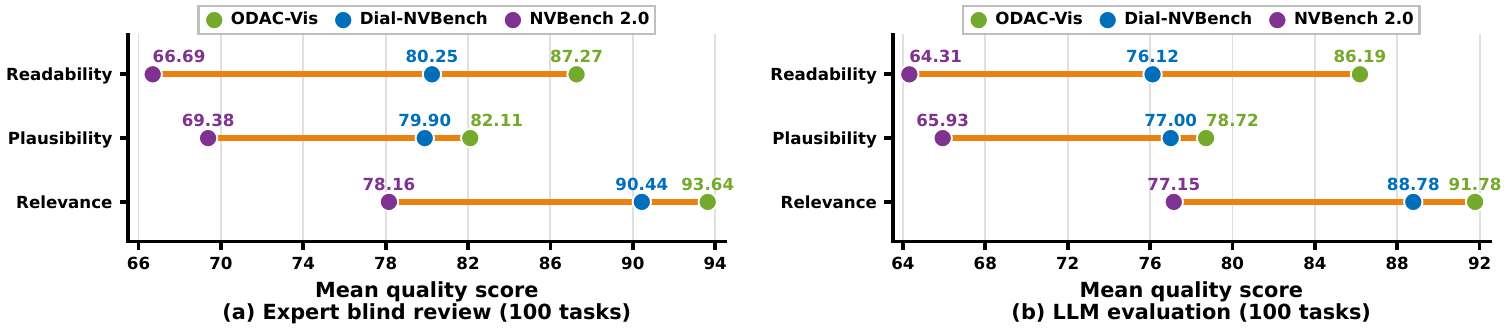}
  \caption{Mean dataset quality scores for \textsc{ODAC-Vis},
    Dial-NVBench, and NVBench~2.0 on 100 randomly sampled tasks per
    benchmark.
    Panel~(a) reports expert blind review; panel~(b) reports LLM
    evaluation with the same three dimensions.}
  \label{fig:dataset-quality}
\end{figure*}

\subsection{Benchmark Statistics}
\label{sec:odac-stats}

Table~\ref{tab:odac-stats} summarizes the released \textsc{ODAC-Vis}
corpus.
The benchmark contains 640 multi-turn tasks over 206 Spider databases
and 139 domains, stratified partitioned into 448 train, 96 validation, and 96 test
instances.
Rather than maximizing instance count alone, the \textsc{ODAC-Vis} prioritizes
\emph{distributional control}: every split preserves balanced scenario
and complexity quotas, validation and test are held to equal size for
stable anchor evaluation, and test domains are disjoint from train and
validation so that success requires locating the right sources rather
than memorizing a fixed schema catalog.
Most tasks remain short enough for interactive analysis, yet the corpus
deliberately retains longer clarify-and-code trajectories and a full
cross-section of typed ambiguities so that models cannot succeed through
single-turn chart guessing alone.

Table~\ref{tab:benchmark-comparison} situates \textsc{ODAC-Vis} against six
representative text-to-visualization and data-access benchmarks.
Existing resources typically isolate one deployment challenge at a time:
single-turn text-to-vis sets assume a known table or database;
conversational benchmarks add dialogue but keep the schema fixed;
ambiguity-aware corpora annotate alternative encodings without requiring
clarification turns; and open-domain text-to-SQL benchmarks such as
TACO~\cite{deng2026taco} evaluate cross-database retrieval and composition
in SQL alone, without executable chart supervision.
\textsc{ODAC-Vis} is designed to stress-test the full analyst workflow in
one benchmark: models must discover target sources from a fully exposed
206-database collection under a domain-held-out split, resolve typed
ambiguities through clarify dialogue, compose evidence across
databases when needed, and emit verified Python programs with reference
charts at every code turn.
To our knowledge, it is the only corpus in the comparison that jointly
supports Python chart generation, open-domain access, cross-domain
evaluation, multi-turn dialogue, verified ambiguity gold, and
cross-database SQL composition.

\begin{table}[t]
  \caption{Scale statistics of \textsc{ODAC-Vis}.}
  \label{tab:odac-stats}
  \centering
  \footnotesize
  \renewcommand{\arraystretch}{1.04}
  \setlength{\tabcolsep}{2pt}
  \begin{tabular}{@{}l@{\hspace{5pt}}r@{\hspace{12pt}}l@{\hspace{5pt}}r@{}}
  \toprule
  \statpanel{Scale and split}
  Tasks & 640 & Tables & 1{,}056 \\
  Train / val / test & 448 / 96 / 96 & Visible databases & 206 \\
  Domains & 139 & Test domains & 15 \\
  \midrule
  \statpanel{Complexity tiers}
  Single-table & \multicolumn{3}{r@{}}{\statrow{192}{30.0}{0.300}} \\
  Single-db join & \multicolumn{3}{r@{}}{\statrow{320}{50.0}{0.500}} \\
  Cross-database & \multicolumn{3}{r@{}}{\statrow{128}{20.0}{0.200}} \\
  \midrule
  \statpanel{Domains used per split}
  Train split & 124 & Val split & 96 \\
  Test split & 15 & Domains per task (med./mean) & 1.0 / 1.2 \\
  \midrule
  \statpanel{Conversation scenarios}
  Drilldown & \multicolumn{3}{r@{}}{\statrow{320}{50.0}{0.500}} \\
  Iterate & \multicolumn{3}{r@{}}{\statrow{320}{50.0}{0.500}} \\
  \midrule
  \statpanel{User--assistant turn count}
  3 turns & \multicolumn{3}{r@{}}{\statrow{270}{42.2}{0.422}} \\
  4 turns & \multicolumn{3}{r@{}}{\statrow{116}{18.1}{0.181}} \\
  5 turns & \multicolumn{3}{r@{}}{\statrow{155}{24.2}{0.242}} \\
  6 turns & \multicolumn{3}{r@{}}{\statrow{36}{5.6}{0.056}} \\
  7 turns & \multicolumn{3}{r@{}}{\statrow{45}{7.0}{0.070}} \\
  8 turns & \multicolumn{3}{r@{}}{\statrow{18}{2.8}{0.028}} \\
  \midrule
  \statpanel{Tasks containing ambiguity type}
  Column selection & \multicolumn{3}{r@{}}{\statrow{120}{18.8}{0.188}} \\
  Aggregation granularity & \multicolumn{3}{r@{}}{\statrow{119}{18.6}{0.186}} \\
  Filter logic & \multicolumn{3}{r@{}}{\statrow{120}{18.8}{0.188}} \\
  Unit scale & \multicolumn{3}{r@{}}{\statrow{120}{18.8}{0.188}} \\
  Chart type & \multicolumn{3}{r@{}}{\statrow{120}{18.8}{0.188}} \\
  Encoding mapping & \multicolumn{3}{r@{}}{\statrow{120}{18.8}{0.188}} \\
  Axis label & \multicolumn{3}{r@{}}{\statrow{120}{18.8}{0.188}} \\
  Interactivity & \multicolumn{3}{r@{}}{\statrow{120}{18.8}{0.188}} \\
  \bottomrule
  \end{tabular}
\end{table}

\subsection{Dataset Quality Analysis}
\label{sec:dataset-quality}

Benchmark comparisons are only meaningful when the underlying supervision
is faithful to the stated task.
We therefore conduct a controlled quality audit of \textsc{ODAC-Vis}
together with two widely used text-to-visualization resources:
Dial-NVBench~\cite{wong2024dialnvbench}, which models multi-turn
visualization dialogue over a fixed given table or schema, and NVBench~2.0~\cite{nvbench22025},
which injects ambiguity into the request and retains multiple
valid chart specifications per request.

Each instance receives three independent scores on a 0--100 scale:
\begin{itemize}
  \item \emph{Relevance}: whether the gold answer targets the analysis
    implied by the request.
  \item \emph{Plausibility}: whether the answer satisfies stated
    constraints and forms a coherent query or chart specification.
  \item \emph{Readability}: presentation quality of the answer text and,
    when applicable, the reference chart.
\end{itemize}

The audit applies this rubric to the gold supervision in each corpus as
follows:
\begin{itemize}
  \item \textsc{ODAC-Vis}: multi-turn clarification and code turns,
    with reference charts produced from verified executable code.
  \item Dial-NVBench: conversational question--answer trajectories
    toward the final visualization goal.
  \item NVBench~2.0: natural-language requests against gold Vega-Lite
    specifications and rendered reference charts when multiple valid
    encodings are annotated.
\end{itemize}

We randomly sample 100 tasks from each benchmark with a fixed seed so
that both evaluators inspect the same instances.
Two experts conduct a blind review using the shared rubric described above, with dataset identity and task provenance concealed.
An LLM evaluator (DeepSeek-V4.1-Flash) applies the same dimensions with structured prompts
and returns a score together with a short rationale for each task.
Both protocols score one instance at a time along a single dimension,
yielding three judgments per task.

Figure~\ref{fig:dataset-quality} summarizes the mean scores.
\textsc{ODAC-Vis} attains the highest quality across all dimensions,
reaching 93.6/82.1/87.3 under expert review and 91.8/78.7/86.2 under
LLM evaluation.
Both evaluation modes agree on the relative ordering of the three
corpora.
The advantage of \textsc{ODAC-Vis} is most visible on readability,
where execution-validated reference charts and typed ambiguity
resolution yield clearer supervision than alternative gold encodings in
NVBench~2.0.
These results support the release of \textsc{ODAC-Vis} as a benchmark
suited to evaluating open-domain, ambiguity-aware,
multi-turn conversational text-to-visualization models.

\subsection{Illustrative Example}
\label{sec:illustrative-example}

\begin{figure}[!htbp]
  \centering
  \includegraphics[width=0.98\columnwidth]{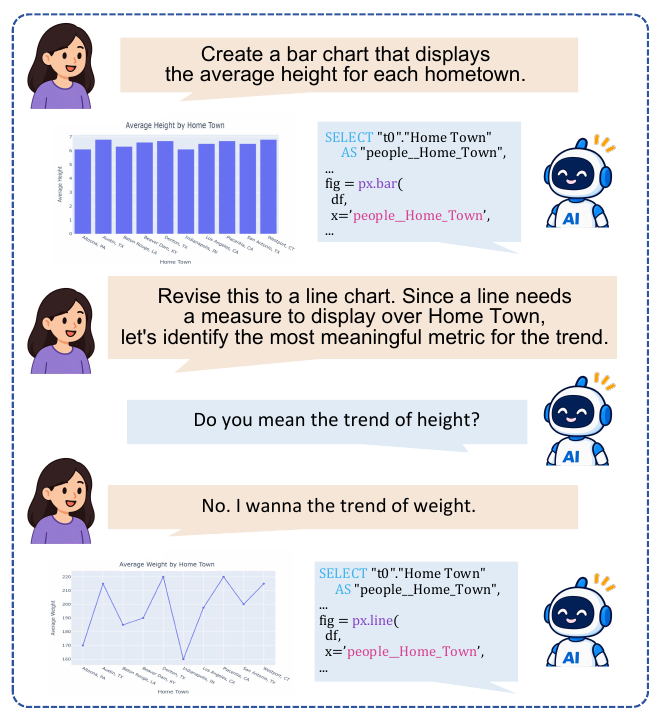}
  \caption{An illustrative task of \textsc{ODAC-Vis}.
    The first request is unambiguous, so the model retrieves the
    relevant data and writes a bar chart.
    A later iteration leaves the measure unspecified; the model asks a
    clarification question and emits the line chart after the user
    selects weight.}
  \label{fig:dataset-case}
\end{figure}

Figure~\ref{fig:dataset-case} shows one task of \textsc{ODAC-Vis}.
The user states a chart request and does not name a
database.
The model first judges whether the request is ambiguous.
The opening request already fixes the analysis: a bar chart of the
average height for each hometown.
Because the measure, the grouping, and the chart type are all specified,
no clarification is required.
The model locates the database that can answer the request, retrieves
the corresponding rows, and writes visualization code that renders the
chart.

Later turns may revise the chart from the previous round
(\emph{iterate}) or ask a deeper question about the same analysis
(\emph{drilldown}).
In this example the user asks to replace the bar chart with a line
chart.
The chart type is fixed, but the measure is not.
The model therefore asks whether the trend of height is intended.
After the user clarifies that the measure should be weight, the model
retrieves that column and renders the line chart.

\section{RSIV-Harness}
\label{sec:rsiv}

\textsc{ODAC-Vis} requires an agent to find the relevant sources in
an exposed database collection.
If the request is ambiguous, the agent must resolve that ambiguity in
dialogue before it returns an executable chart.
\textsc{RSIV-Harness} meets that task with a language model whose
weights stay fixed and a harness that is revised from the agent's own
traces.

\begin{figure*}[t]
  \centering
  \includegraphics[
    width=\textwidth,
    height=0.44\textheight,
    keepaspectratio=false
  ]{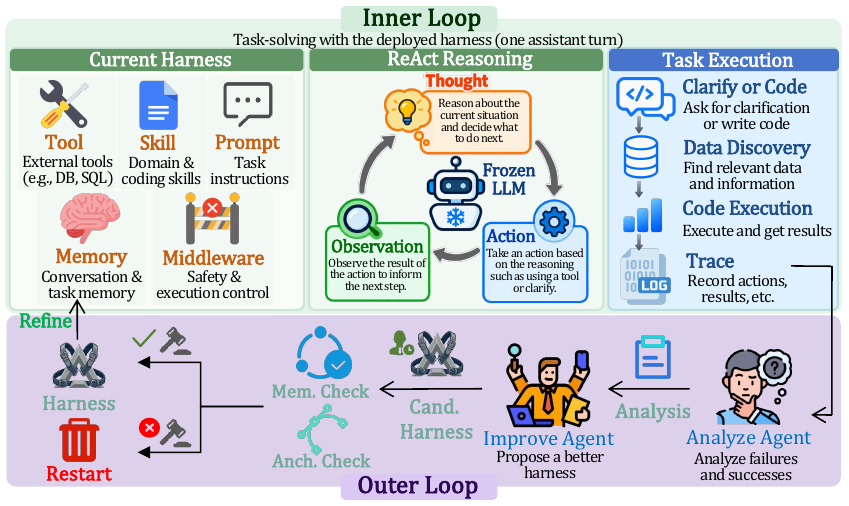}
  \caption{Architecture of \textsc{RSIV-Harness}.
    The inner task-solving loop answers one assistant turn under the
    deployed harness, and the outer self-improvement loop revises
    that harness from traces accumulated across tasks.}
  \label{fig:architecture}
\end{figure*}

This section proceeds from the architecture to the inner loop, then to
the outer self-improvement loop, and finally to the four-layer stability
rules and the theorems that follow from them.

\subsection{Design Philosophy and Architecture Overview}
\label{sec:rsiv-overview}

Existing visualization systems still face a fundamental limit that
blocks practical deployment.
A traditional pipeline fixes its procedure before the first request, so
a failure that keeps recurring can be removed only by rewriting that
procedure or by retraining the
model.
Harness agents improve on this design: a frozen model can choose its
next action from the observation it has just
received.
The harness that governs the choice is nevertheless fixed at
deployment, so the agent stays rigid and has little flexibility once
the scenario grows more complex than the behavior written for it.
To address this limitation, we propose \textsc{RSIV-Harness}, which
keeps the harness as the acting layer of a frozen model and adds
self-improvement that revises the harness from failures accumulated
across tasks.

\textsc{RSIV-Harness} implements a hierarchical,
bi-level recursive self-improvement process, comprising an inner
task-solving loop and an outer self-improvement loop
(Fig.~\ref{fig:architecture}).
The inner loop answers one assistant turn by following the harness that
is currently deployed.
The outer loop reads the traces of many such turns and proposes a
revision of that harness.
A four-layer stability rule framework then accepts or rejects the
revision before it can replace the deployed harness.
The inner loop therefore supplies the evidence, the outer loop turns
that evidence into a candidate, and the rules decide which harness the
next inner loop will use.
Self-improvement starts from a minimal harness, leaves the
outer loop free to choose what to refine and how to refine it, and
uses the refine, restart, anchor check, and memory check rules as the
hard constraints that keep this freedom stable.

\subsection{Inner Task Loop}
\label{sec:rsiv-inner}

The inner loop follows the ReAct pattern~\cite{yao2023react} and incrementally constructs an answer by analyzing the current request, invoking the registered tools, integrating the returned observations, and determining the next step.
The policy is parameterized by the current harness \(H\), while the
model weights stay fixed.
Consider one assistant turn of a task \(x\).
After \(t\) steps the interaction trajectory is
\begin{equation}
\label{eq:trajectory}
h_t = \bigl[(m_i, a_i, o_i)\bigr]_{i=1}^{t},
\end{equation}
where \(m_i\) is a short decision summary, \(a_i\) is the action, and
\(o_i\) is the observation returned by the environment.

The harness defines a policy \(\pi_H\) and an environment operator
\(\mathcal{T}_H\).
Writing \(\bar{h}_t\) for the context shown to the policy, the
next step is
\begin{equation}
\label{eq:policy}
(m_{t+1}, a_{t+1}) = \pi_H(x, \bar{h}_t),
\qquad
o_{t+1} = \mathcal{T}_H(a_{t+1}),
\end{equation}
and the trajectory grows by concatenation,
\begin{equation}
\label{eq:append}
h_{t+1} = h_t \oplus (m_{t+1}, a_{t+1}, o_{t+1}).
\end{equation}
Each step emits one action from a closed set.
The action requests one registered tool, or it terminates the turn
with a clarification.
A tool request names one member of the registered set
\(\mathcal{T}(H)\).

The inner loop stops when a legal clarification or a finished program
has been produced, or when \(m(H)\) policy steps have been used.
Middleware, once a version introduces it, may replace the raw
trajectory by a shorter effective context.
If the latest refresh occurred at step \(\tau \le t\),
\begin{equation}
\label{eq:context}
\bar{h}_t
=
z_{\tau}
\oplus
\bigl[(m_i, a_i, o_i)\bigr]_{i=\tau+1}^{t},
\qquad
z_{\tau} = f_H(h_{\tau}),
\end{equation}
and \(\bar{h}_t = h_t\) when no refresh has occurred.
The refreshed state retains the observations that later steps still
need, including evidence that ruled a source out.
This is the role of autonomous context update, specialized here to schema evidence and
dialogue state.
Until middleware exists, the policy conditions on the raw trajectory.

A clarification ends the turn when it satisfies the dialogue rule.
A chart program is checked and then executed in the sandbox.
A failed execution is kept as the evidence for the next repair.
Let \(e_{<j}\) be the execution failures observed before the \(j\)-th
generation inside the same turn, and define
\begin{equation}
\label{eq:pj}
p_j = \Pr(\mathrm{exec\_ok} \mid e_{<j}).
\end{equation}
The harness allows at most \(K(H)\) repairs after the first generation,
so the turn may execute the program at most \(K(H)+1\) times.
The probability that some attempt is executable is
\begin{equation}
\label{eq:retry}
q\bigl(K(H)\bigr)
=
1 - \prod_{j=0}^{K(H)} (1 - p_j).
\end{equation}
Each repair is conditioned on the concrete error of the previous
attempt and reuses the source evidence already stored in \(h_t\).
The repair rewrites the program attempt and leaves \(H\) as it was.
Section~\ref{sec:rsiv-outer} defines the separate map that revises the
harness.

\subsection{Outer Self-Improvement Loop}
\label{sec:rsiv-outer}

The outer loop treats the harness as the state of a recursive
operator.
Let \(S\) be a training panel drawn once from the train split,
stratified on the factors reported in Section~\ref{sec:exp-dataset},
and then reused for every round.
Let \(V(H)\) be the task success rate of \(\pi_H\) on a fixed
validation anchor.
A task contributes to this rate only when every assistant turn takes
the reference stance and completes, and every gold-code turn both
executes and yields a result table equivalent to the reference table
under the execution-equivalence comparator of~\eqref{eq:exec-acc}.
Because the panel and the anchor stay fixed, a change in \(V\) is
attributable to the harness.

One round begins from the trace of the current harness,
\begin{equation}
\label{eq:evaluate}
\tau(H) = \mathrm{Evaluate}(\pi_H, S).
\end{equation}
Analysis reads \(\tau(H)\) together with the verdict set
\(\mathcal{R}\) of the candidate rejected in the previous round.
It groups traces that break the same part of the task contract, and
only then asks a proposal model to interpret a bounded sample of each
group.
We write the resulting assessment as
\begin{equation}
\label{eq:analyze}
g(H)
=
G\bigl(\tau(H), \mathcal{R}\bigr)
=
\bigl(\mathcal{P}, \mathcal{I}\bigr).
\end{equation}
Here \(\mathcal{P}\) is a bounded memory of protected train records
\((x,t,d,b)\), consisting of a task, turn, metric, and the
baseline score achieved by a committed harness; \(\mathcal{I}\)
is the set of unresolved failure clusters. Records are admitted at
score 1 for stance, executability, routing, and execution accuracy, at
a positive score for clarification quality, and at score 0.75 or above
for chart structure.
The proposal is
\begin{equation}
\label{eq:improve}
H' = \mathrm{Improve}\bigl(g(H), H\bigr),
\end{equation}
and it contains at most \(E\) edits.
Each edit names the component it changes and the failure evidence it
is meant to address.
A legal proposal is then tested by the stability rules of
Section~\ref{sec:rsiv-control}.

The candidate is then submitted to the commit rule of
Section~\ref{sec:rsiv-control}.
Let \(H^\star\) be the committed harness with the best anchor score, and
let \(B\) be that score.
Write \(Q\) for the stratified regression sample defined below, and
abbreviate the commit predicate by
\begin{equation}
\label{eq:accept}
\mathrm{Accept}(H')
=
\mathbf{1}\bigl[
\mathrm{MC}(H') = 1
\land
\mathrm{AC}(H') = 1
\bigr],
\end{equation}
where \(\mathrm{MC}\) is the memory check and \(\mathrm{AC}\) is the
anchor check defined in Section~\ref{sec:rsiv-control}.
The recursive operator is
\begin{equation}
\label{eq:phi}
H_{n+1}
=
\Phi(H_n)
=
\begin{cases}
H'_n & \text{if } \mathrm{Accept}(H'_n) = 1, \\
H^\star_n & \text{otherwise.}
\end{cases}
\end{equation}
The seed is \(H_0\).
On acceptance, eligible protected records from the new version are
folded into the bounded memory after the deployed harness has been evaluated again on
\(S\),
\begin{equation}
\label{eq:refresh}
\mathcal{P}_{n+1}
=
\mathrm{Refresh}_C\bigl(\mathcal{P}_n, \mathcal{A}(H_{n+1})\bigr).
\end{equation}
Here \(\mathcal{A}(H)\) is the set of eligible protected records
observed under \(H\), and \(C\) is the memory capacity.
The next analysis can then attend to the clusters that remain in
\(\mathcal{I}\).
On rejection, \(\Phi\) restores \(H^\star_n\).
The candidate is archived with its verdict, and that verdict becomes
\(\mathcal{R}\) for the following round.
A subsequent acceptance clears \(\mathcal{R}\).
Rejection is the harness-level restart rule: the failed revision is withdrawn, and the reason
for the failure is retained.
Acceptance is the refine rule: verified protected records
stay in \(\mathcal{P}\), and the next round focuses on what
\(\mathcal{I}\) still contains.
The champion score \(B\) is replaced only when the accepted anchor
score is strictly larger, so \(B\) never decreases.
A commit inside the factor \((1-\varepsilon)\) still become the
current harness.

Algorithm~\ref{alg:evolve} records the order in which these steps are
executed.
The seed is evaluated on \(S\) before any candidate is proposed, and
the bounded protected-record memory is filled from that evaluation.
The seed is also scored on the anchor, which initializes \(B\) and
\(H^\star\).
After every commit or rejection, the harness left in place is evaluated
again on \(S\).
That new trace is the input to the next application of \(\Phi\).
New protected records enter the regression memory only from this
post-decision evaluation, and only when the round committed.

\begin{algorithm}[t]
\caption{Recursive self-improvement of the harness.}
\label{alg:evolve}
\small
\begin{algorithmic}[1]
\Require seed \(H_0\), panel \(S\), anchor \(S_{\mathrm{val}}\)
\Require tolerances \(\varepsilon,\delta\), memory capacity \(C\), edit budget \(E\), rounds \(N\)
\State Evaluate \(\pi_{H_0}\) on \(S\) and seed \(\mathcal{P}\) with \(\mathcal{A}(H_0)\)
\State \(B \gets V(H_0)\) on \(S_{\mathrm{val}}\);
\(H^\star \gets H_0\); \(\mathcal{R} \gets \emptyset\)
\For{\(n = 0,1,\ldots,N-1\)}
  \State \(g(H_n) \gets G(\tau(H_n), \mathcal{R})\)
  \State \(H' \gets \mathrm{Improve}(g(H_n), H_n)\)
    \Comment{at most \(E\) edits}
  \State Draw a stratified sample \(Q\) from \(\mathcal{P}\)
  \If{\(\mathrm{MC}(H') = 0\)}
    \State Reject \(H'\); \(H_{n+1} \gets H^\star\)
    \State \(\mathcal{R} \gets \{\text{verdict of } H'\}\)
  \ElsIf{\(\mathrm{AC}(H') = 0\)}
    \State Reject \(H'\); \(H_{n+1} \gets H^\star\)
    \State \(\mathcal{R} \gets \{\text{verdict of } H'\}\)
  \Else
    \State Commit \(H_{n+1} \gets H'\) and clear \(\mathcal{R}\)
    \If{\(V(H') > B\)}
      \State \(B \gets V(H')\); \(H^\star \gets H'\)
    \EndIf
  \EndIf
  \State \(\tau(H_{n+1}) \gets \mathrm{Evaluate}(\pi_{H_{n+1}}, S)\)
  \If{\(H_{n+1}\) was committed}
    \State Absorb \(\mathcal{A}(H_{n+1})\) into \(\mathcal{P}\) up to capacity \(C\)
  \EndIf
\EndFor
\Return{\(H^\star\)}
\end{algorithmic}
\end{algorithm}

\subsection{Stability Rule-Enhanced Evolution}
\label{sec:rsiv-control}

One contribution of the method is to keep self-improvement at its greatest freedom while using rules to guarantee the stability of that evolution.
We propose a four-layer stability rule framework and provide the theoretical guarantees that follow from it.

\subsubsection{Four-Layer Stability Rule}
\label{sec:rsiv-rules}

The framework has four layers, written \(\mathcal{R} = \{RF, RS, AC, MC\}\).
A candidate that passes the anchor check and the memory check is refined into the deployed harness.
Every other candidate, by contrast, restarts the deployed harness from the current champion.

\paragraph{Refine rules (RF): protected progress.}
An accepted candidate stays deployed, its eligible protected records refresh the bounded memory, and the next round attends to the clusters that are still open.

\subparagraph{\textbf{(i) RF-Rule 1}: Protected-record refresh.}
Eligible records of the accepted harness are added to the bounded memory.
Define the refresh map used in~\eqref{eq:refresh} by
\begin{equation}
\label{eq:rl-refresh}
\mathrm{Refresh}_C(\mathcal{P}, A)
=
\mathrm{Cap}_C\bigl(\mathrm{Merge}(\mathcal{P}, A)\bigr),
\end{equation}
where \(\mathrm{Merge}\) keys records by task, turn, and dimension and
retains the stored baseline for an existing key, while
\(\mathrm{Cap}_C\) applies the deterministic capacity-\(C\) retention
policy.
On \(\mathrm{Accept}(H'_n) = 1\),
\begin{equation}
\label{eq:rl-apply}
\mathcal{P}_{n+1}
=
\mathrm{Refresh}_C\bigl(\mathcal{P}_n, \mathcal{A}(H_{n+1})\bigr),
\end{equation}
where \(\mathcal{A}(H)\) is the set of eligible turn--metric records observed under the given harness \(H\).

\subparagraph{\textbf{(ii) RF-Rule 2}: Residual focus.}
The next assessment reads the protected-record memory and the clusters that remain unresolved.
\begin{equation}
\label{eq:rl-focus}
G\bigl(\tau(H_{n+1}), \emptyset\bigr)
=
\bigl(\mathcal{P}_{n+1}, \mathcal{I}_{n+1}\bigr),
\end{equation}
where \(\tau(H_{n+1})\) is the trace of the deployed harness on the training panel and \(\mathcal{I}_{n+1}\) is the set of failure clusters still open after~\eqref{eq:rl-apply}.

\subparagraph{\textbf{(iii) RF-Rule 3}: Feedback reset.}
Acceptance clears the rejection verdict that the previous round stored.
\begin{equation}
\label{eq:rl-clear}
\mathcal{R}_{n+1}
=
\emptyset
\qquad\text{if}\qquad
\mathrm{Accept}(H'_n) = 1,
\end{equation}
where \(\mathcal{R}_{n+1}\) is the verdict set supplied to the next call of \(G\).

\paragraph{Restart rules (RS).}
A rejected candidate is withdrawn from deployment.
Its rejection reason is retained for the next analysis.

\subparagraph{\textbf{(i) RS-Rule 1}: Champion restoration.}
Rejection returns the deployed harness to the champion.
\begin{equation}
\label{eq:rs-restore}
H_{n+1}
=
H_n^\star
\qquad\text{if}\qquad
\mathrm{Accept}(H'_n) = 0,
\end{equation}
where \(H_n^\star\) is the committed harness whose anchor score equals the champion value \(B_n\).

\subparagraph{\textbf{(ii) RS-Rule 2}: Verdict retention.}
The failed check is passed to the next assessment.
\begin{equation}
\label{eq:rs-verdict}
\mathcal{R}_{n+1}
=
\{v(H'_n)\}
\qquad\text{if}\qquad
\mathrm{Accept}(H'_n) = 0,
\end{equation}
where \(v(H')\) records which predicate failed and which sampled records regressed.

\subparagraph{\textbf{(iii) RS-Rule 3}: Frozen memory.}
A rejected candidate adds no solved task.
\begin{equation}
\label{eq:rs-frozen}
\mathcal{P}_{n+1}
=
\mathcal{P}_n
\qquad\text{if}\qquad
\mathrm{Accept}(H'_n) = 0.
\end{equation}

\paragraph{Anchor check rules (AC): validation floor.}
A candidate may be refined only when its score on the fixed validation anchor stays close to the best committed score.

\subparagraph{\textbf{(i) AC-Rule 1}: Relative admission.}
The anchor test compares the candidate with the current champion.
\begin{equation}
\label{eq:ac}
\mathrm{AC}(H')
=
\mathbf{1}\bigl[
V(H') \ge (1 - \varepsilon)\, B
\bigr],
\end{equation}
where \(V(H)\) is the task success rate of \(\pi_H\) on the fixed anchor, \(B\) is the champion score, and \(\varepsilon \in [0, 1]\) is the tolerance.
Let \(j \in \{1,\ldots,\ell\}\) index the bounded remediation cycles
within one outer round, where \(\ell\) is the maximum number of
candidate cycles. An anchor failure returns control to Analyze with its
verdict, and becomes a terminal rejection only in the last cycle:
\begin{equation}
\label{eq:ac-remediate}
\mathrm{AC}(H'_{n,j})=0
\implies
\begin{cases}
g_{n,j+1}=G\bigl(\tau(H_n^\star),\{v_{\mathrm{AC}}(H'_{n,j})\}\bigr),
& j<\ell,\\
\mathrm{Reject}_{\mathrm{AC}}(n), & j=\ell.
\end{cases}
\end{equation}

\subparagraph{\textbf{(ii) AC-Rule 2}: Strict champion update.}
The champion score increases only when an accepted candidate strictly exceeds it.
\begin{equation}
\label{eq:bupdate}
B_n
=
\begin{cases}
V(H'_n)
&
\text{if }\mathrm{Accept}(H'_n) = 1\text{ and }V(H'_n) > B_{n-1},\\
B_{n-1}
&
\text{otherwise.}
\end{cases}
\end{equation}
Thus \(H^\star\) changes only in the first branch, while \(B_n \ge B_{n-1}\).

\subparagraph{\textbf{(iii) AC-Rule 3}: Band deployment.}
An accepted candidate becomes the current harness even when it does not set a new champion.
\begin{equation}
\label{eq:ac-band}
H_{n+1}
=
H'_n
\qquad\text{whenever}\qquad
\mathrm{Accept}(H'_n) = 1,
\end{equation}
whereas \(H_n^\star\) is updated only by~\eqref{eq:bupdate}.
A later restart therefore returns directly to the champion, which may itself differ from the harness deployed inside the band.

\paragraph{Memory check rules (MC): protected records.}
The regression memory blocks a commit that degrades too many sampled turn--metric achievements of committed harnesses.

\subparagraph{\textbf{(i) MC-Rule 1}: Stratified sample.}
The sample cycles through the groups represented in the memory.
Let \(\Gamma(\mathcal{P})\) be the set of all nonempty triples of metric dimension, dialogue scenario, and complexity tier represented in \(\mathcal{P}\).
\begin{equation}
\label{eq:cycle}
Q
=
\mathrm{Cycle}\bigl(\mathcal{P}, \Gamma(\mathcal{P})\bigr),
\end{equation}
where \(\mathrm{Cycle}\) takes one record from each group in turn until the sample budget is filled.
A group is visited again only after every nonempty group has contributed.

\subparagraph{\textbf{(ii) MC-Rule 2}: Regression bound.}
A sampled record \(r=(x,t,d,b)\) regresses when the candidate's score
\(s_d(H';x,t)\) falls below its stored baseline \(b\).
\begin{equation}
\label{eq:reg}
\mathrm{Reg}_Q(H')
=
\sum_{(x,t,d,b) \in Q}
\mathbf{1}\bigl[
s_d(H';x,t) < b
\bigr],
\end{equation}
\begin{equation}
\label{eq:mc}
\mathrm{MC}(H')
=
\mathbf{1}\bigl[
\mathrm{Reg}_Q(H') \le \delta
\bigr],
\end{equation}
where \(\delta\) is a nonnegative integer tolerance.
The predicate \(\mathrm{MC}\) is the memory conjunct of~\eqref{eq:accept}.
A memory failure, for its part, follows the same bounded return to Analyze, with the
regressed records included in its verdict:
\begin{equation}
\label{eq:mc-remediate}
\mathrm{MC}(H'_{n,j})=0
\implies
\begin{cases}
g_{n,j+1}=G\bigl(\tau(H_n^\star),\{v_{\mathrm{MC}}(H'_{n,j})\}\bigr),
& j<\ell,\\
\mathrm{Reject}_{\mathrm{MC}}(n), & j=\ell.
\end{cases}
\end{equation}
Thus neither check terminates the round after its first failure. The
round is rejected only if all \(\ell\) remediation cycles fail.

\subparagraph{\textbf{(iii) MC-Rule 3}: Seeded growth.}
The memory exists before any candidate is proposed, and it is refreshed only through a refine step.
\begin{equation}
\label{eq:mc-seed}
\mathcal{P}_0
=
\mathrm{Cap}_C\bigl(\mathcal{A}(H_0)\bigr),
\end{equation}
and \(\mathcal{P}_{n+1}\) follows~\eqref{eq:rl-apply} after acceptance and~\eqref{eq:rs-frozen} after rejection.

\subsubsection{Theoretical Guarantees}
\label{sec:rsiv-theory}

The stability rules constrain which evolutionary update may be deployed.
Let an evolution context be
\begin{equation*}
\mathcal{C}
=
(H, \mathcal{P}, \mathcal{I}, B, \mathcal{R}),
\end{equation*}
where \(H\) is the current harness, \(\mathcal{P}\) the bounded protected-record memory, \(\mathcal{I}\) the open clusters, \(B\) the champion score, and \(\mathcal{R} = \{RF, RS, AC, MC\}\) the rule set.
Throughout, \(V\) takes values in \([0, 1]\).

\paragraph{Formal theorems.}
The framework yields four guarantees.

\subparagraph{\textbf{(i) Theorem 1} (Protected-record refresh).}
\textit{If a candidate is accepted, the bounded memory is refreshed with eligible records from the deployed harness and the stored rejection is cleared:}
\begin{equation}
\label{eq:thm-refine}
\mathrm{Accept}(H'_n) = 1
\implies
\mathcal{P}_{n+1}
=
\mathrm{Refresh}_C\bigl(\mathcal{P}_n, \mathcal{A}(H_{n+1})\bigr)
\;\land\;
\mathcal{R}_{n+1}
=
\emptyset.
\end{equation}

\textbf{Proof sketch.}
RF-Rule 1 gives~\eqref{eq:rl-apply}.
RF-Rule 3 gives \(\mathcal{R}_{n+1} = \emptyset\).
RF-Rule 2 then instantiates the next assessment as \(G(\tau(H_{n+1}), \emptyset) = (\mathcal{P}_{n+1}, \mathcal{I}_{n+1})\), so the open set passed forward is \(\mathcal{I}_{n+1}\).

\subparagraph{\textbf{(ii) Theorem 2} (Restart invariance).}
\textit{If a candidate is rejected, deployment returns to the champion, the protected-record memory is unchanged, and the rejection verdict is retained:}
\begin{equation}
\label{eq:thm-restart}
\mathrm{Accept}(H'_n) = 0
\implies
H_{n+1} = H_n^\star
\;\land\;
\mathcal{P}_{n+1} = \mathcal{P}_n
\;\land\;
\mathcal{R}_{n+1} = \{v(H'_n)\}.
\end{equation}

\textbf{Proof sketch.}
RS-Rule 1 is~\eqref{eq:rs-restore}, hence \(H_{n+1} = H_n^\star\).
RS-Rule 3 is~\eqref{eq:rs-frozen}, hence \(\mathcal{P}_{n+1} = \mathcal{P}_n\).
RS-Rule 2 is~\eqref{eq:rs-verdict}, hence \(\mathcal{R}_{n+1} = \{v(H'_n)\}\).
The next call is therefore \(G(\tau(H_n^\star), \{v(H'_n)\})\).

\subparagraph{\textbf{(iii) Theorem 3} (Anchor monotonicity).}
\textit{The champion score is nondecreasing, and the harness left in deployment stays above the anchor floor of that score:}
\begin{equation}
\label{eq:thm-anchor}
B_n \ge B_{n-1}
\qquad\text{and}\qquad
V(\tilde{H}_n) \ge (1 - \varepsilon)\, B_n,
\end{equation}
where \(\tilde{H}_n\) denotes the harness deployed at the end of round \(n\).

\textbf{Proof sketch.}
From~\eqref{eq:bupdate}, the two branches are \(B_n = V(H'_n) > B_{n-1}\) and \(B_n = B_{n-1}\), so \(B_n \ge B_{n-1}\).
Since \(V \in [0, 1]\), the sequence \((B_n)\) is bounded.
Let \(\mathrm{Accept}(H'_n) = 1\).
AC-Rule 1 yields \(V(H'_n) \ge (1 - \varepsilon) B_{n-1}\), and~\eqref{eq:ac-band} sets \(\tilde{H}_n = H'_n\).
If \(V(H'_n) > B_{n-1}\), then \(B_n = V(H'_n)\) and \(V(\tilde{H}_n) = B_n \ge (1 - \varepsilon) B_n\).
If \(V(H'_n) \le B_{n-1}\), then \(B_n = B_{n-1}\) and \(V(\tilde{H}_n) \ge (1 - \varepsilon) B_n\).
Let \(\mathrm{Accept}(H'_n) = 0\).
RS-Rule 1 sets \(\tilde{H}_n = H_n^\star\) with \(V(H_n^\star) = B_n\), hence \(V(\tilde{H}_n) = B_n \ge (1 - \varepsilon) B_n\).

\subparagraph{\textbf{(iv) Theorem 4} (Anti-oscillation).}
\textit{Acceptance implies that the stratified regression count does not exceed the memory tolerance:}
\begin{equation}
\label{eq:thm-memory}
\mathrm{Accept}(H') = 1
\implies
\mathrm{Reg}_Q(H') \le \delta.
\end{equation}

\textbf{Proof sketch.}
Expand~\eqref{eq:accept} to \(\mathrm{MC}(H') = 1 \land \mathrm{AC}(H') = 1\).
Substitute~\eqref{eq:mc} to obtain \(\mathrm{Reg}_Q(H') \le \delta\).
By~\eqref{eq:cycle}, \(Q = \mathrm{Cycle}(\mathcal{P}, \Gamma(\mathcal{P}))\) emits one element of each \(\gamma \in \Gamma(\mathcal{P})\) before repeating any group.
Therefore, whenever \(|\Gamma(\mathcal{P})| > 1\), the count in~\eqref{eq:reg} is not computed on a single metric--scenario--tier group.
MC-Rule 3 initializes \(\mathcal{P}_0 = \mathrm{Cap}_C(\mathcal{A}(H_0))\) before the first proposal, so the same bound applies to the protected seed records retained in memory.

\section{Experiments}
\label{sec:experiments}

\subsection{Experimental Setup}
\label{sec:exp-setup}

\subsubsection{Dataset}
\label{sec:exp-dataset}

All experiments use \textsc{ODAC-Vis}.
We use the train split to drive the outer loop of
\textsc{RSIV-Harness}. The stratified training panel \(S\)
contains 48 tasks from train only domains. Their failure traces are
analyzed to propose harness edits. The anchor check evaluates every
candidate on a panel of 48 validation tasks.
Each panel contains one task for every combination of the three
difficulty tiers, the eight ambiguity types, and the two conversation
scenarios.
All of the test split is reserved exclusively for final evaluation.
It comprises 96 tasks over 15 held-out domains that never appear in
train or validation, so reported scores measure open-domain source
discovery, ambiguity handling, and multi-turn chart generation rather
than memorization of a fixed schema catalog.
All baselines and evolved harness variants are evaluated on this test
split under model history: each turn conditions on the model's own prior
assistant outputs within the same task.

\subsubsection{Baselines}
\label{sec:exp-baselines}

We compare \textsc{RSIV-Harness} against four progressively more sophisticated baselines: direct prompting, static
harnesses, ReAct, and MultiVis-Agent:
\begin{itemize}
  \item \textbf{Direct Prompt.} A single LLM call per turn with no tool
  access. The model receives the full schema catalog together with the conversation history and NLQ of this turn.
  \item \textbf{Static Harnesses.} The minimal seed harness \(H_0\) of
  \textsc{RSIV-Harness}, held fixed and never evolved.
  A seed already fitted to \textsc{ODAC-Vis} would contaminate the
  attribution of every later edit, since a gain could no longer be
  separated from the seed.
  The minimal seed forces every component the outer loop adds to earn
  its place against measured traces.
  \item \textbf{ReAct~\cite{yao2023react}.} An ReAct
  baseline equipped with carefully designed, task-specific tools. At each turn, it runs a thought-action-observation loop.
  \item \textbf{MultiVis-Agent~\cite{lu2026multivis}.} A state-of-the-art multi-agent system for
  text-to-visualization, adapted to \textsc{ODAC-Vis} task with minimal modifications.
  \item \textbf{RSIV-Harness (Ours).}
  Our Proposed method.
  An inner ReAct loop clarifies the request or generates
  a visualization program, while an outer loop improves the harness
  from traces under a four-layer stability rule framework.
\end{itemize}

\subsubsection{Evaluation Metrics}
\label{sec:exp-metrics}

To evaluate visualization generation systems, particularly those handling open-domain, ambiguity-aware,
multi-turn conversational tasks like \textsc{RSIV-Harness}, we adopt the following metrics.
Let \(\mathcal{T}\) be the assistant turns in the evaluation split.
For turn \(t\), \(y_t\) is the gold stance and \(\hat{y}_t\) is the predicted stance.
Write \(\mathcal{C}=\{t\in\mathcal{T}: y_t=\mathrm{code}\}\) and
\(\mathcal{Q}=\{t\in\mathcal{T}: y_t=\mathrm{clarify}\}\).
\begin{itemize}
  \item \textbf{Executability} tests whether the emitted program can be
    run to completion.
    It isolates basic viability before asking whether the analysis is
    correct:
    \begin{equation}
      \label{eq:executability}
      \mathrm{Exec}
      =
      \frac{100}{|\mathcal{C}|}
      \sum_{t\in\mathcal{C}}
      \mathbf{1}\!\bigl[\mathrm{Run}(\hat{p}_t)\bigr],
    \end{equation}
    where \(\hat{p}_t\) is the predicted program and
    \(\mathrm{Run}(\hat{p}_t)=1\) when a valid predicted program runs to
    completion.
  \item \textbf{Database Routing Accuracy} checks whether the program
    accesses exactly the target databases for the task.
    In an open-domain setting, retrieving the wrong source can yield a
    fluent but irrelevant chart even when the code runs successfully.
    Let \(D(\hat{p}_t)\) be the set of databases accessed by the prediction, and
    \(D_t^{\star}\) the gold target set:
    \begin{equation}
      \label{eq:routing}
      \mathrm{Route}
      =
      \frac{100}{|\mathcal{C}|}
      \sum_{t\in\mathcal{C}}
      \mathbf{1}\!\bigl[D(\hat{p}_t)=D_t^{\star}\bigr].
    \end{equation}
  \item \textbf{Stance Accuracy} measures whether an agent chooses the
    appropriate response mode---clarification or executable
    code---at each assistant turn.
    Because downstream dialogue and prediction depend on this
    decision, stance errors can invalidate an otherwise plausible
    program or chart:
    \begin{equation}
      \label{eq:stance}
      \mathrm{Stance}
      =
      \frac{100}{|\mathcal{T}|}
      \sum_{t\in\mathcal{T}}
      \mathbf{1}\!\bigl[\hat{y}_t=y_t\bigr].
    \end{equation}
  \item \textbf{Execution Accuracy} compares the tabular results pro-
  duced by the ground-truth and predicted programs, determining whether the pre-
  dicted SQL returns the same result as the ground-truth
  SQL:
    \begin{equation}
      \label{eq:exec-acc}
      \mathrm{ExAcc}
      =
      \frac{100}{|\mathcal{C}|}
      \sum_{t\in\mathcal{C}}
      \mathbf{1}\!\bigl[
        \exists\,\pi:\; \pi(\hat{R}_t)\equiv R_t^{\star}
      \bigr],
    \end{equation}
    where \(R_t^{\star}\) and \(\hat{R}_t\) are the gold and predicted
    tables.
  \item \textbf{Clarification Quality} evaluates how well predicted
  clarification questions recover the intended ambiguity resolution.
  A completed clarification is compared with the
  gold clarification and any annotated alternatives.
  For each \(n\in\{1,2,3,4\}\), modified \(n\)-gram precision is
  \begin{equation}
    \label{eq:clarify-bleu4}
    p_n
    =
    \frac{
      \sum_{g\in \mathrm{Gram}_n(h)}
      \min\!\bigl(c_h(g),\, c_{\mathcal{R}}^{\max}(g)\bigr)
    }{
      \sum_{g\in \mathrm{Gram}_n(h)} c_h(g)
    },
  \end{equation}
  where \(h\) is the predicted clarification, \(\mathcal{R}\) is the
  reference set, \(c_h(g)\) counts occurrences of \(g\) in \(h\), and
  \(c_{\mathcal{R}}^{\max}(g)\) is the maximum reference count of \(g\); otherwise
  \begin{equation}
    \label{eq:clarify-bleu4-geo}
    \mathrm{BLEU\text{-}4}(h,\mathcal{R})
    =
    100 \cdot
    \exp\!\Bigl(\tfrac{1}{4}\sum_{n=1}^{4}\log p_n\Bigr),
  \end{equation}
  without a brevity penalty.
  The reported score is the mean turn-level BLEU-4,
  \begin{equation}
    \label{eq:clarify-agg}
    \mathrm{Clarify}
    =
    \frac{1}{|\mathcal{Q}|}
    \sum_{t\in\mathcal{Q}}
    s_t,
  \end{equation}
  where \(s_t=\mathrm{BLEU\text{-}4}(h_t,\mathcal{R}_t)\) when the
  prediction clarifies and \(s_t=0\) otherwise.
  \item \textbf{Chart Structure} compares the figure produced by the
    predicted program with the gold figure.
    Each figure is reduced to a signature
    \((M, X, Y, c, a_x, a_y, I, R)\):
    \(M\) is the mark types;
    \(X\) and \(Y\) are the per-trace role sequences of the \(x\) and
    \(y\) encodings;
    \(c\) holds when the figure has multiple traces or a legend group;
    \(a_x\) and \(a_y\) are the axis titles;
    \(I\) is the set of interaction kinds;
    and \(R\) is the multiset of plotted rows.
    \begin{equation}
      \label{eq:chart-struct}
      \begin{aligned}
      \mathrm{Struct}
      &=
      \frac{1}{8}\bigl(
        \mathbf{1}[M=\hat{M}]
        + \mathbf{1}[X=\hat{X}] \\
      &\quad
        + \mathbf{1}[Y=\hat{Y}]
        + \mathbf{1}[c=\hat{c}]
        + \alpha_x + \alpha_y \\
      &\quad
        + J(I,\hat{I})
        + \mathbf{1}[R\equiv\hat{R}]
      \bigr),
      \end{aligned}
    \end{equation}
    where \(\alpha_x=\mathrm{Title}(a_x,\hat{a}_x)\) and
    \(\alpha_y=\mathrm{Title}(a_y,\hat{a}_y)\).
    The reported score averages the turn values,
    \begin{equation}
      \label{eq:chart}
      \mathrm{Chart}
      =
      \frac{100}{|\mathcal{C}|}
      \sum
      \mathrm{Struct}
    \end{equation}
\end{itemize}

\begin{table}[t]
  \caption{Overall results on the \textsc{ODAC-Vis} test split across different methods.}
  \label{tab:overall-results}
  \centering
  \scriptsize
  \renewcommand{\arraystretch}{1.05}
  \setlength{\tabcolsep}{2pt}
  \resizebox{\columnwidth}{!}{%
  \renewcommand{\multirowsetup}{\centering}
  \newcommand{\methodcell}[1]{\multirow{2}{*}[-2pt]{\makecell[c]{#1}}}
  \sisetup{detect-weight, mode=text, table-format=2.2}
  \begin{tabular}{@{}l l *{6}{S} @{}}
  \toprule
  {Method} & {Model} &
  {Executability} & {Route.} & {Stance} & {Exec Acc.} &
  {Clarify} & {Chart} \\
  \midrule
  \methodcell{Direct\\Prompt} &
    DeepSeek &
    81.32 & 57.59 & 62.25 & \best 31.91 & 4.32 & 52.23 \\
  \cmidrule{2-8}
  & GPT &
    63.04 & 44.36 & 61.00 & 24.51 & 3.02 & 47.46 \\
  \cmidrule{1-8}
  \methodcell{Static\\Harnesses} &
    DeepSeek &
    64.98 & 48.25 & 51.25 & 24.51 & 3.72 & 41.46 \\
  \cmidrule{2-8}
  & GPT &
    51.75 & 30.35 & 54.00 & 26.07 & 2.02 & 42.01 \\
  \cmidrule{1-8}
  \multirow{2}{*}[-2pt]{ReAct} &
    DeepSeek &
    79.38 & 57.59 & 64.25 & 27.24 & 3.93 & 49.57 \\
  \cmidrule{2-8}
  & GPT &
    65.76 & 44.75 & 61.75 & 21.40 & 2.27 & 50.37 \\
  \cmidrule{1-8}
  \methodcell{MultiVis-\\Agent} &
    DeepSeek &
    58.75 & 56.81 & 56.25 & 26.85 & 0.60 & 44.08 \\
  \cmidrule{2-8}
  & GPT &
    53.70 & 50.97 & 58.25 & 25.29 & 0.97 & 40.39 \\
  \cmidrule{1-8}
  \methodcell{\textsc{RSIV-}\\Harness} &
    DeepSeek &
    \best 93.00 & \best 62.26 & \best 67.00 & 31.52 & \best 5.15 & \best 75.47 \\
  \cmidrule{2-8}
  & GPT &
    61.09 & 49.42 & 61.50 & 25.29 & 3.74 & 50.36 \\
  \bottomrule
  \end{tabular}%
  }
\end{table}

\subsubsection{Implementation Details}
\label{sec:exp-details}
We first perform one self-improvement round of \textsc{RSIV-Harness} with
DeepSeek-V4.1-Flash. After this round, we freeze the selected evolved
harness and use the same unchanged harness to evaluate the test split
with DeepSeek-V4.1-Flash and GPT-6 Luna~\cite{openai2026gpt6luna}.
No further evolution or backbone specific adaptation is performed for
GPT-6 Luna. All models used by \textsc{RSIV-Harness} and the baselines
use their default hyperparameter settings, including temperature and
top-p.

\begin{figure*}[t]
  \centering
  \captionsetup[subfigure]{labelformat=parens, labelsep=space, font=scriptsize, skip=2pt}
  \begin{subfigure}[t]{0.155\textwidth}
    \centering
    \includegraphics[width=\linewidth]{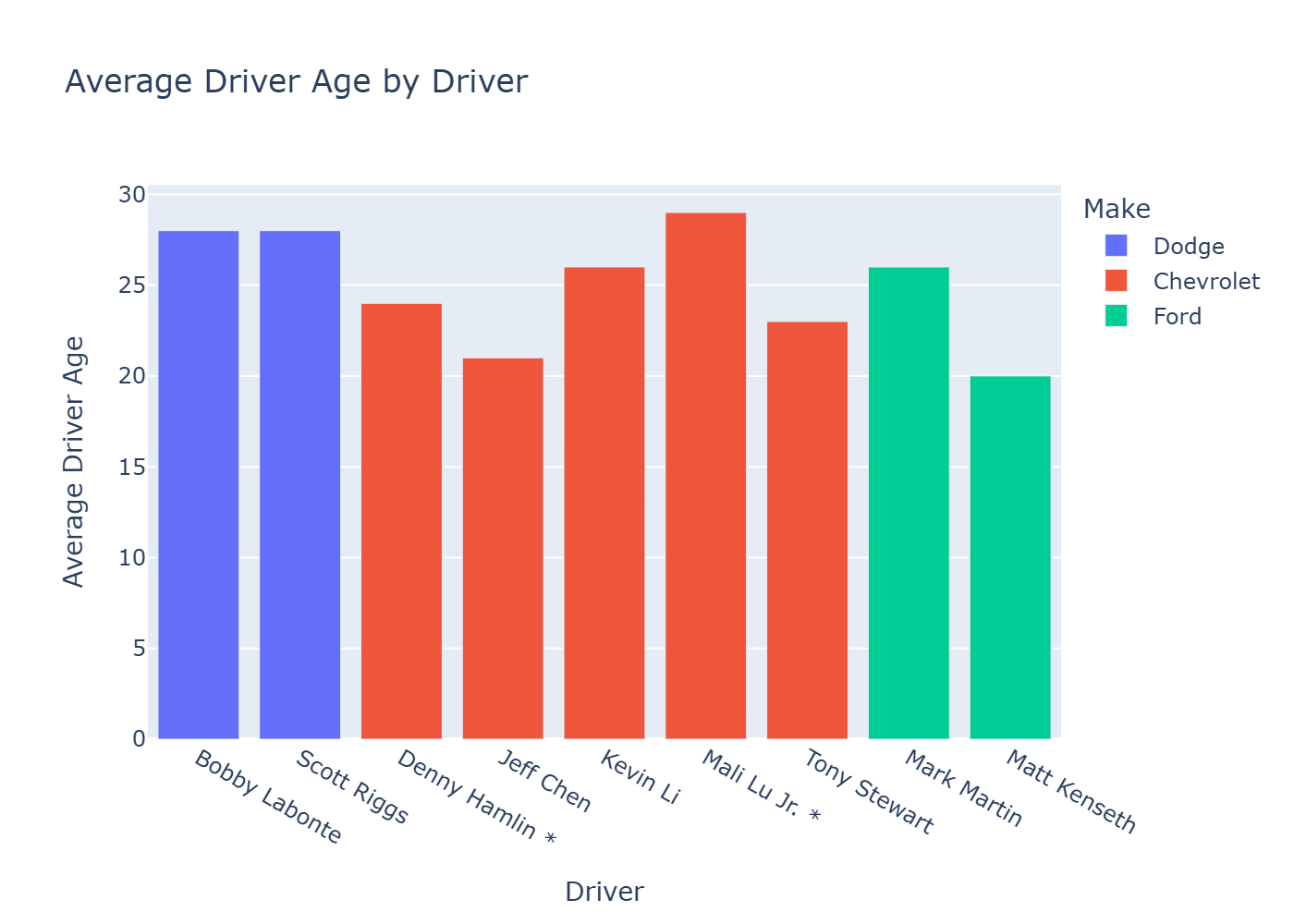}
    \caption{Ground Truth}
    \label{fig:case-gt}
  \end{subfigure}\hfill
  \begin{subfigure}[t]{0.155\textwidth}
    \centering
    \includegraphics[width=\linewidth]{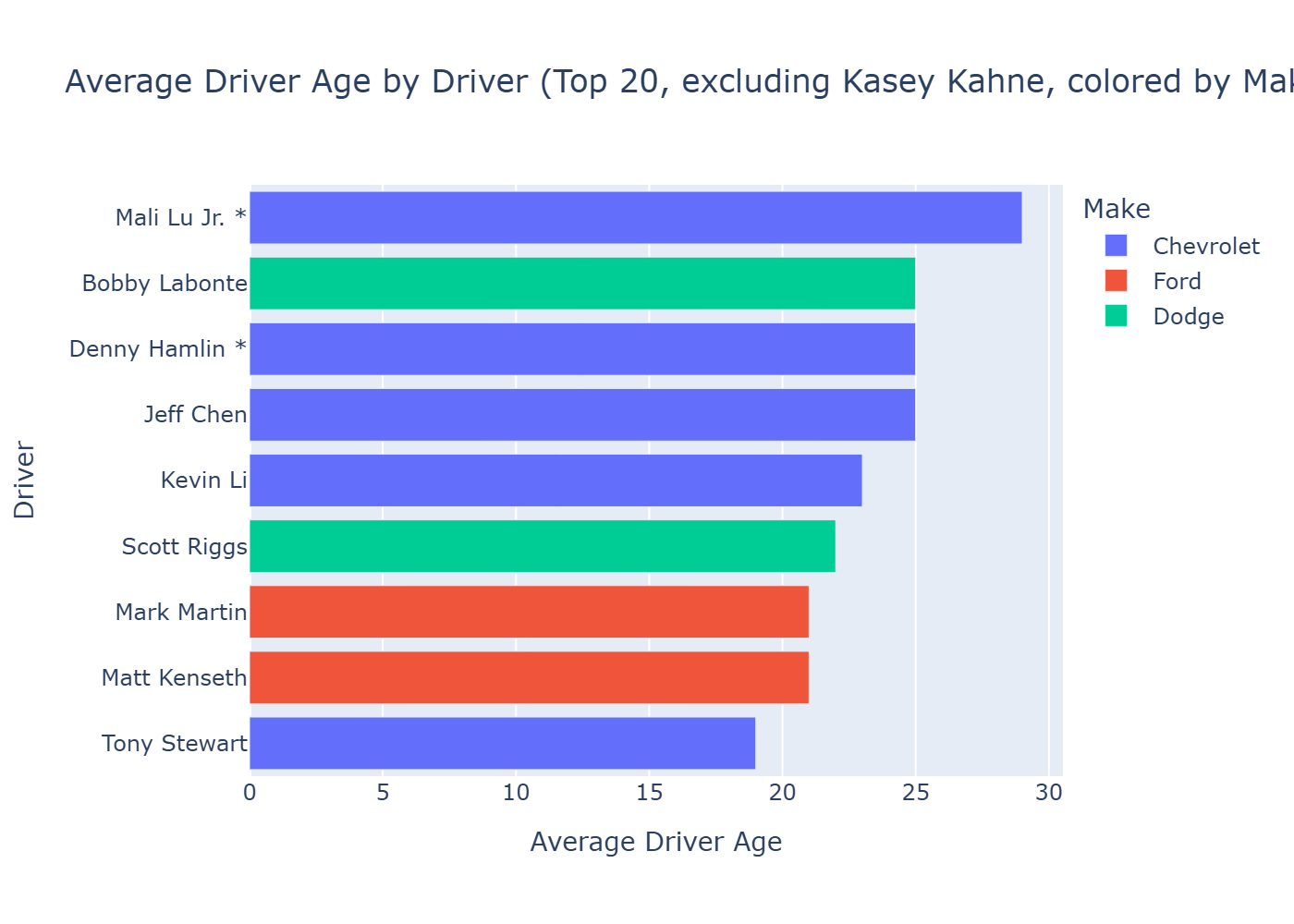}
    \caption{Direct Prompt}
    \label{fig:case-direct}
  \end{subfigure}\hfill
  \begin{subfigure}[t]{0.155\textwidth}
    \centering
    \includegraphics[width=\linewidth]{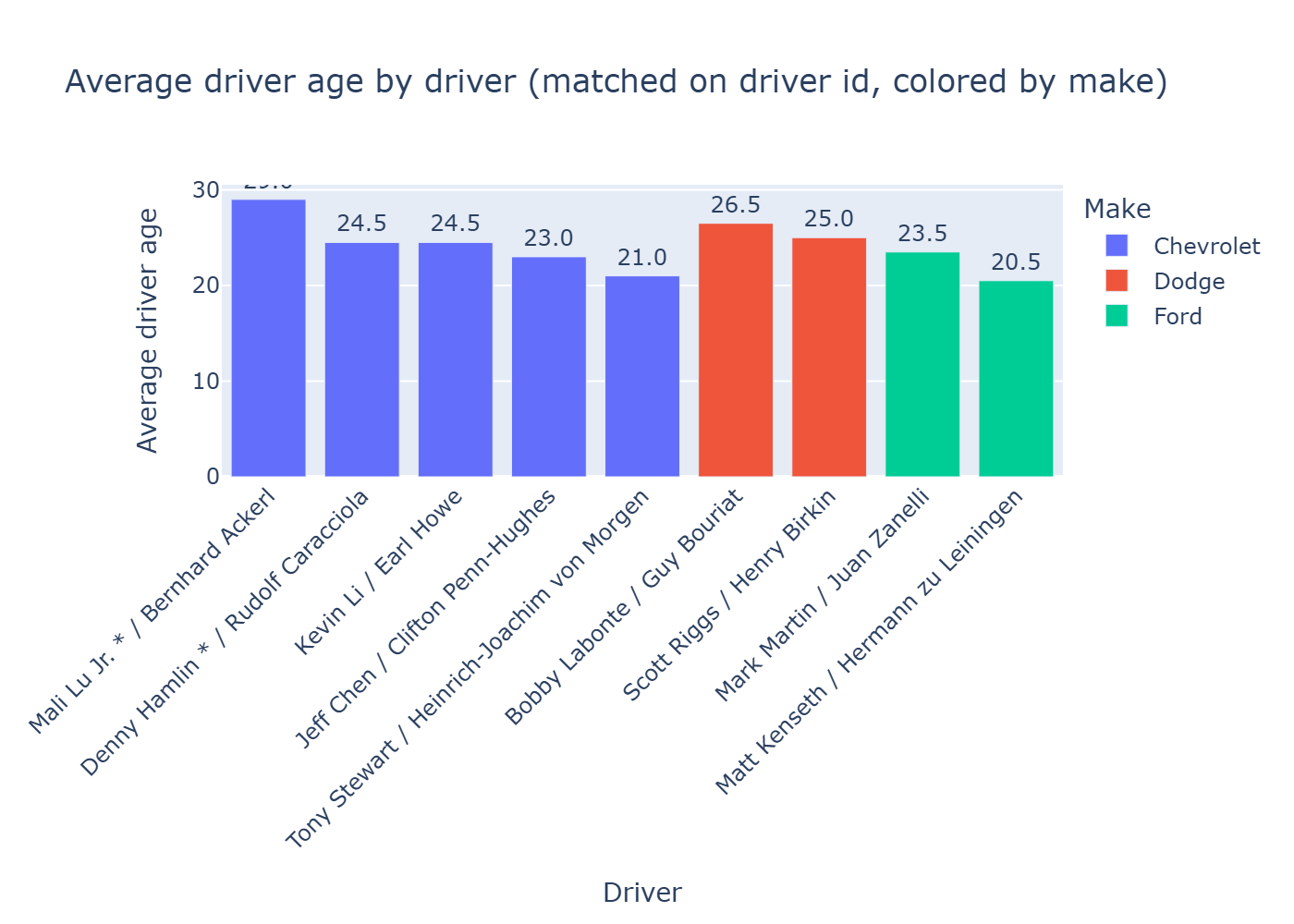}
    \caption{\(H_0\)}
    \label{fig:case-h0}
  \end{subfigure}\hfill
  \begin{subfigure}[t]{0.155\textwidth}
    \centering
    \includegraphics[width=\linewidth]{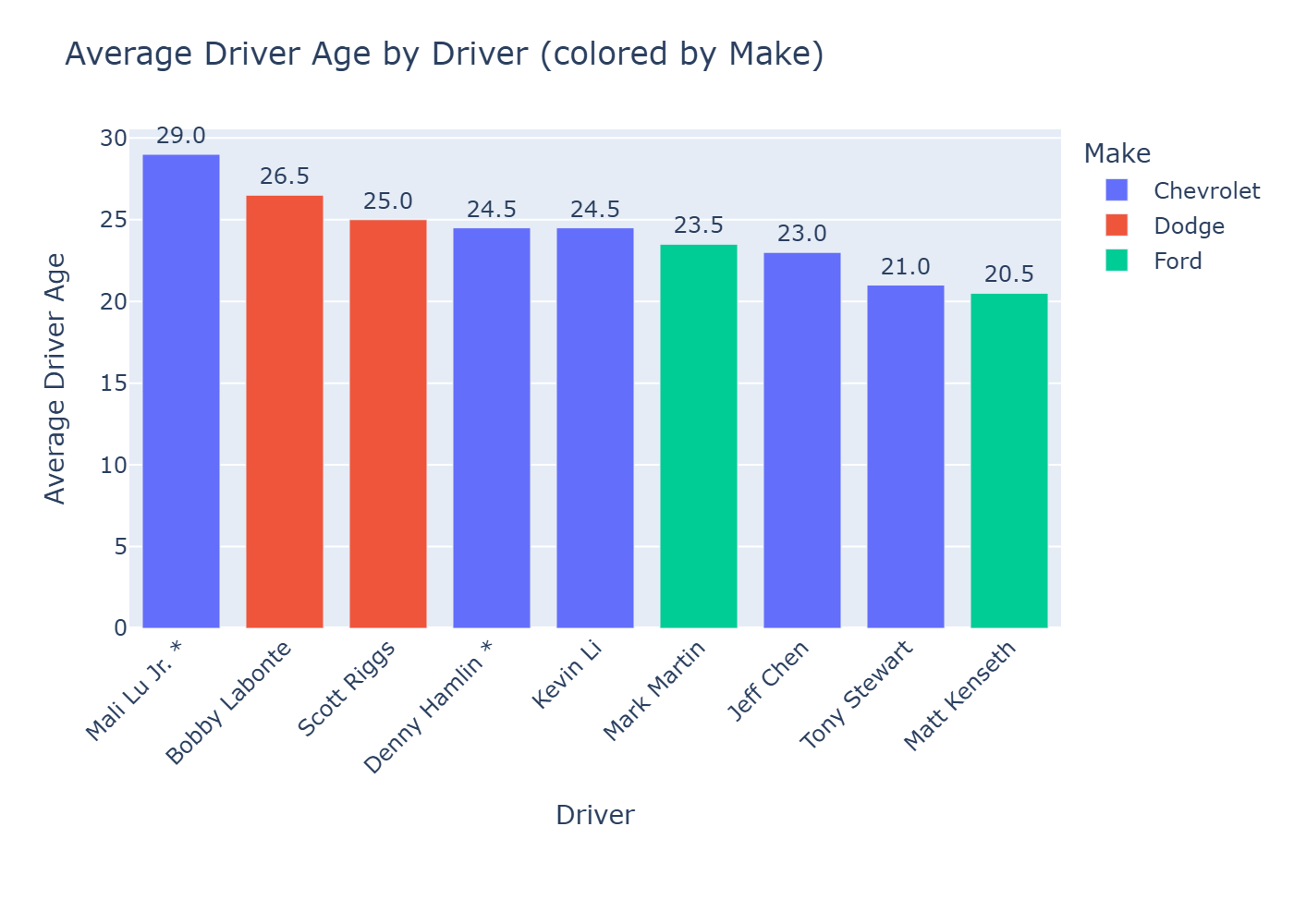}
    \caption{MultiVis-Agent}
    \label{fig:case-multivis}
  \end{subfigure}\hfill
  \begin{subfigure}[t]{0.155\textwidth}
    \centering
    \includegraphics[width=\linewidth]{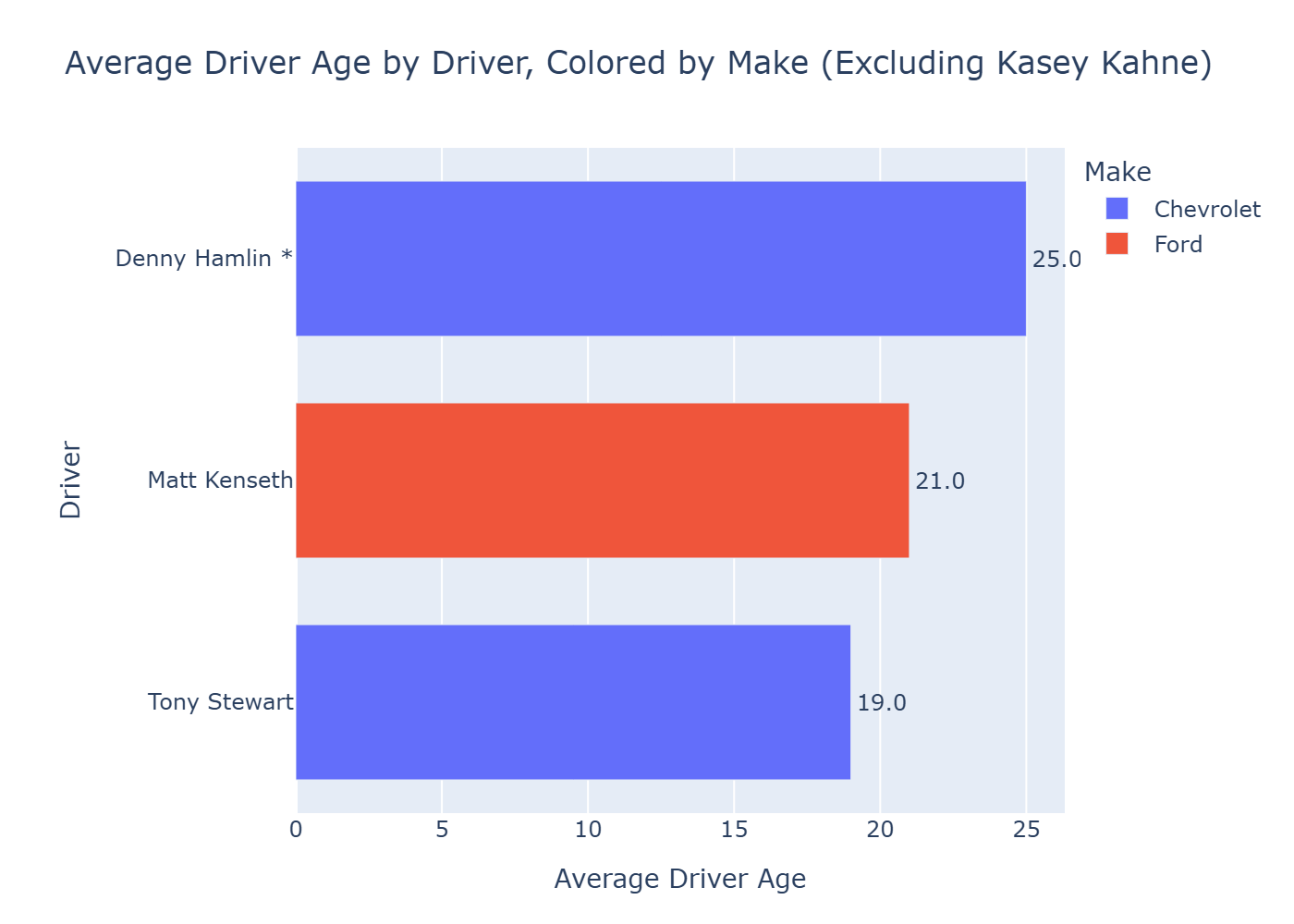}
    \caption{ReAct}
    \label{fig:case-react}
  \end{subfigure}\hfill
  \begin{subfigure}[t]{0.155\textwidth}
    \centering
    \includegraphics[width=\linewidth]{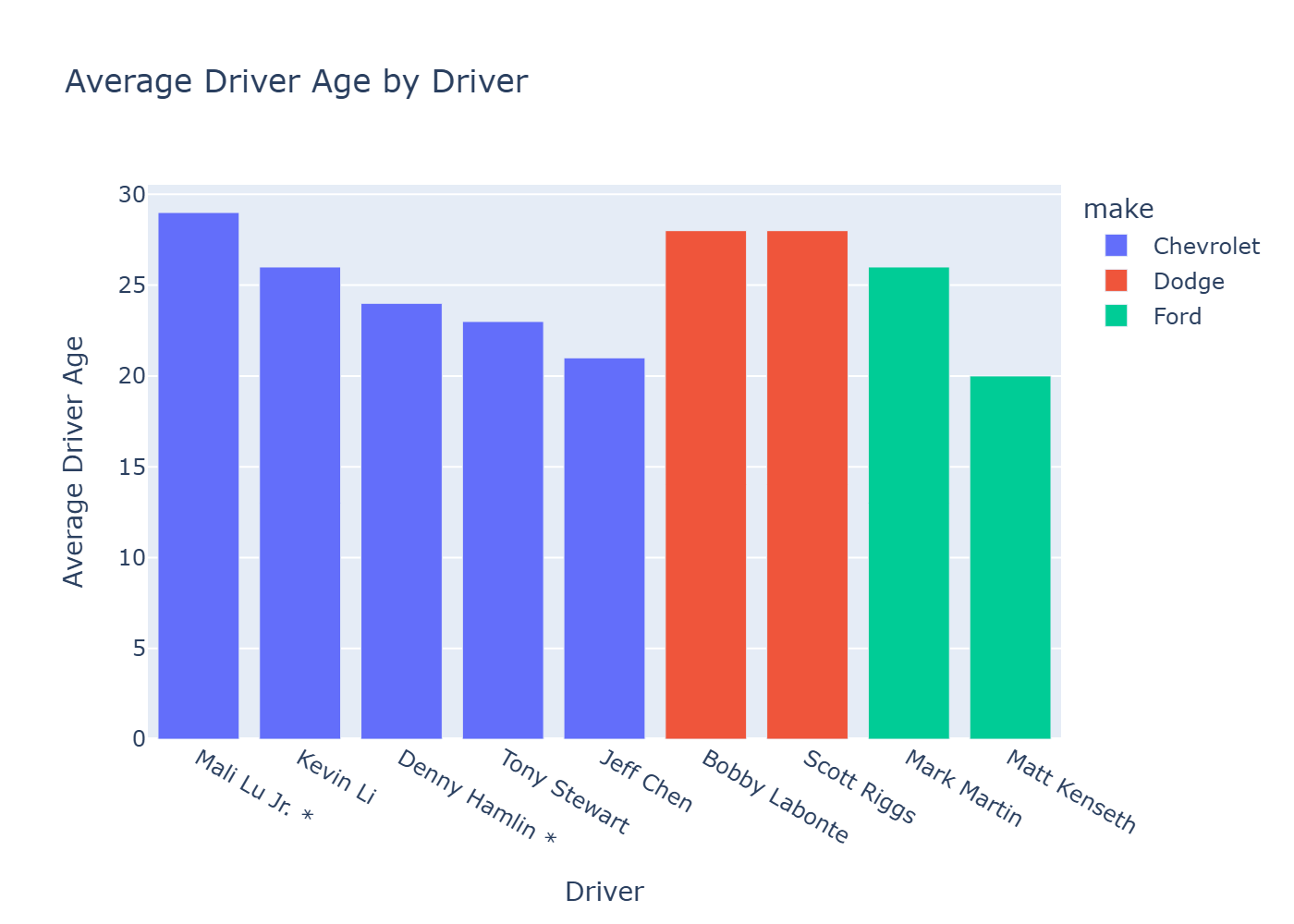}
    \caption{\textsc{RSIV-Harness}}
    \label{fig:case-rsi}
  \end{subfigure}
  \caption{Case study on a \textsc{ODAC-Vis} task.
    The user asks for average driver age by driver, and chooses color by make.
    Direct Prompt and ReAct change the plotted ages and drivers.
    \(H_0\) concatenates names from the two sources, and MultiVis-Agent changes the bar heights.
    \textsc{RSIV-Harness} successfully matches user requirements.}
  \label{fig:case-study}
\end{figure*}

\subsection{Results and Analysis}
\label{sec:exp-results}

\subsubsection{Overall Results}
\label{sec:exp-overall-results}

Table~\ref{tab:overall-results} shows that \textsc{RSIV-Harness} obtains
the strongest DeepSeek results on five of the six metrics. Relative to
the best competing method in each column, it improves executability by
11.68 points, database routing by 4.67 points, stance accuracy by 2.75
points, clarification quality by 0.83 points, and chart structure by
23.24 points. Its execution accuracy of 31.52 is only 0.39 points below
Direct Prompt. The comparison with the static harness isolates the
effect of evolution more directly. Recursive improvement raises the
six metrics by 28.02, 14.01, 15.75, 7.01, 1.43, and 34.01 points in
table order. These gains show that adapting the harness around fixed
model weights improves not only program viability, but also source
selection, dialogue decisions, and the structure of the final chart.

\subsubsection{Impact of LLM Backbone}
\label{sec:exp-backbone}

The GPT results test whether a harness evolved only with DeepSeek can
transfer to another backbone without further optimization. Compared
with the static GPT harness, the frozen evolved harness improves five
of six metrics. The gains are 9.34 points in executability, 19.07 in
database routing, 7.50 in stance accuracy, 1.72 in clarification
quality, and 8.35 in chart structure. Execution accuracy is 0.78 points
lower. The transferred system is nevertheless slightly below the
strongest GPT baseline on most individual metrics. The deficits are
4.67 points for executability, 1.55 for routing, 0.25 for stance, 0.78
for execution accuracy, and 0.01 for chart structure. It achieves the
best clarification score at 3.74. These results demonstrate the applicability of \textsc{RSIV-Harness}
across open-source and proprietary LLM backbones,
with harness improvements learned from one model remaining transferable to
another without adaptation,
although some performance gains remain dependent on the model used during evolution.

\subsection{Ablation Study}
\label{sec:ablation}

\begin{table}[t]
  \caption{Ablation of \textsc{RSIV-Harness} on the \textsc{ODAC-Vis} test split.
  The first variant removes the stability rules. The remaining variants add one evolved component to \(H_0\).}
  \label{tab:ablation}
  \centering
  \scriptsize
  \renewcommand{\arraystretch}{1.05}
  \setlength{\tabcolsep}{2.5pt}
  \resizebox{\columnwidth}{!}{%
  \sisetup{detect-weight, mode=text, table-format=2.2}
  \begin{tabular}{@{}l *{6}{S} @{}}
    \toprule
    {Method} &
    {Exec.} & {Route.} & {Stance} & {Exec Acc.} & {Clarify} & {Chart} \\
    \midrule
    \textsc{RSIV-Harness} &
      \best 93.00 & \best 62.26 & \best 67.00 & \best 31.52 & \best 5.15 & \best 75.47 \\
    \quad - w/o. rule &
      83.27 & 59.14 & 64.50 & 28.02 & 2.41 & 65.94 \\
    \cmidrule{1-7}
    H0 &
      51.36 & 48.25 & 51.25 & 24.51 & 3.72 & 41.46 \\
    \quad + prompt only &
      87.94 & \best 62.26 & 64.00 & 29.96 & \best 3.76 & 71.13 \\
    \quad + tool only &
      84.82 & \best 62.26 & 65.25 & 27.63 & 2.84 & 66.67 \\
    \quad + memory only &
      85.99 & 59.92 & 64.00 & 26.85 & 2.56 & 67.69 \\
    \quad + skill only &
      85.99 & 60.70 & 64.00 & 29.18 & 2.29 & 69.47 \\
    \quad + middleware only &
      \best 90.66 & 56.03 & \best 66.50 & \best 30.35 & 2.07 & \best 72.41 \\
    \bottomrule
  \end{tabular}%
  }
\end{table}

\subsubsection{Stability Rule Analysis}
\label{sec:exp-stability-rule-analysis}

To isolate the effect of stability rules, we repeat one round of harness evolution with
DeepSeek-V4.1-Flash after removing the stability rule framework, keeping all other settings unchanged.
The resulting harness is then frozen and evaluated on the test split using the same model.
As shown in Table~\ref{tab:ablation}, removing the stability rules leads to performance degradation across all six metrics:
executability drops by 9.73 points, database routing by 3.12,
stance accuracy by 2.50, execution accuracy by 3.50, clarification quality by 2.74,
and chart structure by 9.53.
The largest declines occur in executability and chart structure, while the reductions in routing, stance, and clarification quality indicate that the framework contributes to multiple aspects of conversational visualization rather than merely filtering invalid programs.
These results highlight the importance of stability rules in guiding harness evolution, helping preserve previously successful behaviors while enabling improvements from accumulated task failures.

\subsubsection{Harness Component Analysis}
\label{sec:exp-harness-component-analysis}

To assess component contributions, we replace each component of the \(H_0\)
with its evolved counterpart—prompt, tool, memory, skill, or middleware—while keeping
the others unchanged. Table~\ref{tab:ablation} reveals complementary strengths.
The evolved prompt improves routing (62.26), clarification (3.76), and chart structure (71.13),
while the tool interface supports routing (62.26) and stance accuracy (65.25).
Memory benefits routing (59.92) and chart structure (67.69),
whereas skills improve execution accuracy (29.18) and chart structure (69.47).
Middleware achieves the best single-component results in executability (90.66),
stance accuracy (66.50), execution accuracy (30.35), and chart structure (72.41),
highlighting the role of control flow and execution feedback.
The complete harness still exceeds every
single component variant in five metrics.
The combined result therefore arises from coordination among
components rather than from one isolated edit.

\subsection{Case Study}
\label{sec:case-study}
We present a cross-database visualization task involving two car-racing databases joined by driver ID
to demonstrate the capabilities of \textsc{RSIV-Harness}.
The user first requests a bar chart of average driver age,
then excludes Kasey Kahne and uses color to encode vehicle make.
The reference chart (Figure~\ref{fig:case-gt}) therefore contains nine drivers across three makes.
Direct Prompt and ReAct distort the ages and chart orientation, with ReAct omitting six drivers.
\(H_0\) incorrectly concatenates driver names across sources.
MultiVis-Agent keeps the nine names but shifts the ages, for example drawing Bobby Labonte at 26.5 rather than 28.
In contrast, \textsc{RSIV-Harness} preserves the correct drivers, ages, and make-based encoding, closely matching the reference chart.
This demonstrates its ability to integrate cross-database evidence while respecting multi-turn refinements.

\section{Conclusion}
\label{sec:conclusion}

We presented \textsc{ODAC-Vis} and \textsc{RSIV-Harness} as a benchmark
and method for open domain conversational text to visualization.
\textsc{ODAC-Vis} evaluates source discovery, ambiguity resolution, and
multi turn chart refinement under held out domains.
\textsc{RSIV-Harness} improves the harness around a fixed language model through recursive analysis of
failure traces. Its stability rules provide
explicit guarantees for protected record refresh, champion
restoration, anchor monotonicity, and bounded regression.
On the test split, a single round of recursive self-improvement improves
the DeepSeek-based harness across all six evaluation metrics. The same frozen harness also
transfers competitively to GPT-6 Luna. Ablations further show that the
stability rules improve all reported metrics and that the harness
components make complementary contributions. These findings establish
harness evolution as a practical path for improving conversational
visualization agents.

\bibliographystyle{ACM-Reference-Format}
\bibliography{sample-base}

\end{document}